\documentclass{ecai} 

\usepackage{latexsym}
\usepackage{amssymb}
\usepackage{amsmath}
\usepackage{amsthm}
\usepackage{booktabs}
\usepackage{enumitem}
\usepackage{graphicx}
\usepackage{color}
\usepackage{etoc} 

\usepackage{multicol}
\usepackage{multirow}
\usepackage{subfigure}
\usepackage{mathtools}
\usepackage[capitalize,noabbrev]{cleveref}
\usepackage{wrapfig}
\usepackage{algorithmic}
\usepackage{algorithm}
\usepackage{minitoc}

\renewcommand \thepart{}
\renewcommand \partname{}

\newcommand{\BibTeX}{B\kern-.05em{\sc i\kern-.025em b}\kern-.08em\TeX}

\begin{document}

\begin{frontmatter}


\paperid{3235} 

\title{Beyond Pixels: Enhancing LIME with Hierarchical Features and Segmentation Foundation Models}

\author[A]{\fnms{Patrick}~\snm{Knab}}
\author[A,B]{\fnms{Sascha}~\snm{Marton}}
\author[A]{\fnms{Christian}~\snm{Bartelt}} 

\address[A]{Clausthal University of Technology}
\address[B]{University of Mannheim}

\begin{abstract}
LIME (Local Interpretable Model-agnostic Explanations) is a well-known XAI framework for unraveling decision-making processes in vision machine-learning models. 
The technique utilizes image segmentation methods to identify fixed regions for calculating feature importance scores as explanations. Therefore, poor segmentation can weaken the explanation and reduce the importance of segments, ultimately affecting the overall clarity of interpretation.
To address these challenges, we introduce the \textbf{DSEG-LIME} (Data-Driven Segmentation LIME) framework, featuring: \textbf{\textit{i})} a \textit{data-driven} segmentation for \textit{human-recognized} feature generation by \textit{foundation model} integration, and \textbf{\textit{ii})} a user-steered granularity in the \textit{hierarchical segmentation} procedure through \textit{composition}.
Our findings demonstrate that DSEG outperforms on several XAI metrics on pre-trained ImageNet models and improves the alignment of explanations with human-recognized concepts.
\end{abstract}

\end{frontmatter}

\section{Introduction}\label{sec1}
\textbf{Why should we trust you?} The integration of AI-powered services into everyday scenarios, with or without the need for specific domain knowledge, is becoming increasingly common.
Consider autonomous driving or surveillance, where accurate visual recognition of images or video streams is critical. 
In such high-stakes applications, precision and alignment with expert judgment and safety standards are paramount. 
To ensure reliability and safety, stakeholders frequently evaluate AI performance after deployment. 
For example, one might assess whether an AI correctly identifies pedestrians or hazardous objects in real-time footage, accurately detecting potential threats to avoid accidents or breaches in security.
The derived question - "Why should we \textit{trust} the model?" - is a direct reference to \textit{Local Interpretable Model-agnostic Explanations} (LIME) \citep{lime}. LIME seeks to demystify AI decision-making by identifying key features that influence the output of a model, underlying the importance of the Explainable AI (XAI) research domain, particularly when deploying opaque models in real-world scenarios \citep{XAI_concepts, Garreau2021WhatDL, e23010018}.

\begin{figure}[tb]
  \centering
  \includegraphics[width = 0.85\columnwidth]{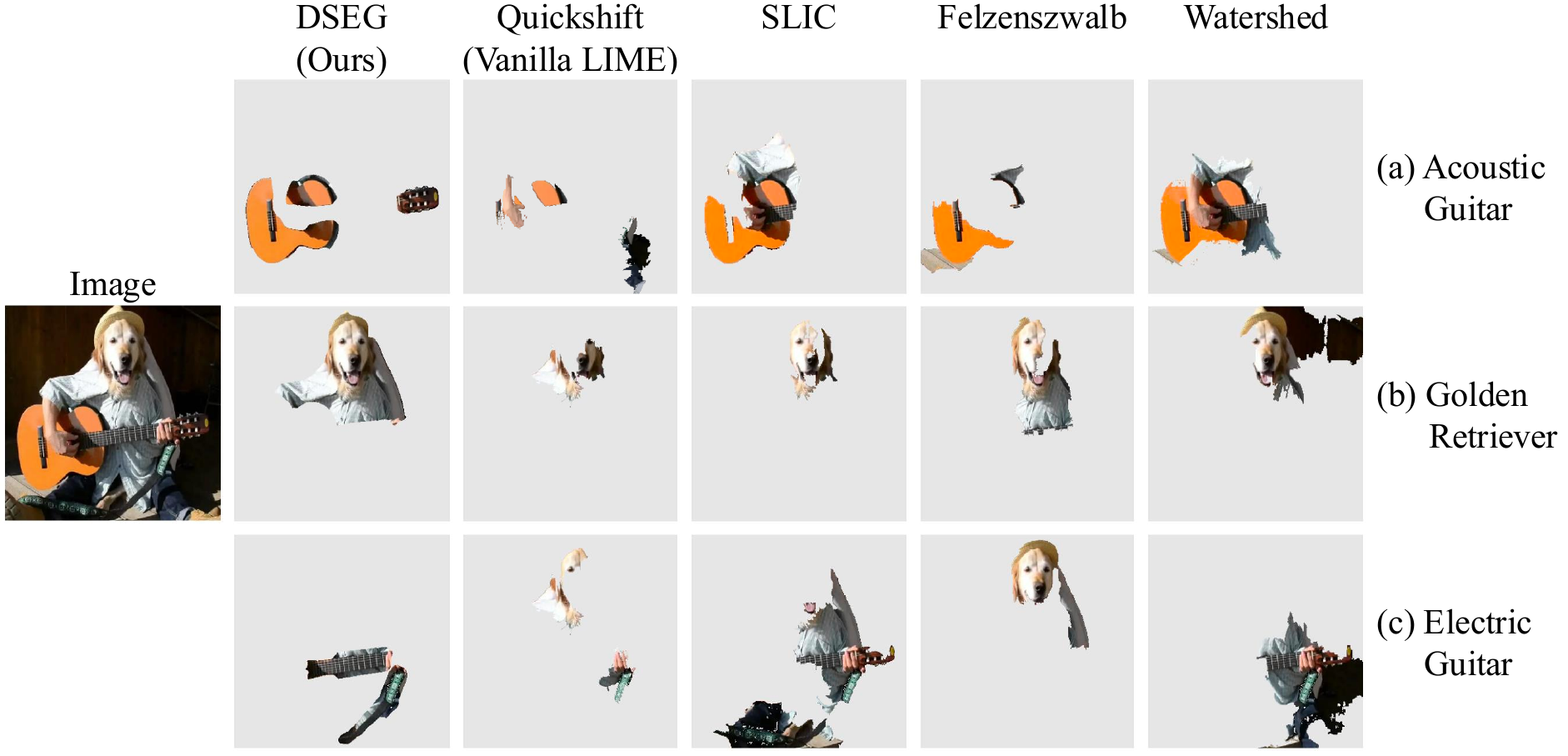}
\caption{\textbf{DSEG vs. other segmentation techniques within LIME.} LIME-generated explanations \citep{lime} for EfficientNetB4 \citep{tan2019efficientnet} using various segmentation methods: DSEG (ours) with SAM \citep{kirillov2023segment}, Quickshift \citep{10.5555/3454287.3454867}, SLIC \citep{6205760}, Felzenszwalb \citep{Felzenszwalb2004}, and Watershed \citep{6976891}. The top predictions are ‘Acoustic Guitar’ ($p$ = 0.31), ‘Golden Retriever’ ($p$ = 0.24), and ‘Electric Guitar’ ($p$ = 0.07). }
\label{fig:dog}
\end{figure}

\textbf{Ambiguous explanations.} \label{problem} 
LIME relies on conventional superpixel segmentation (e.g., graph‑ or clustering‑based methods \citep{wang2017superpixel}) to isolate image regions for classification explanations. 
However, these regions often lack semantic coherence or compositional structure, such as distinguishing car doors, wheels, or headlights, which undermines both clarity and relevance. 
Because LIME defaults to these arbitrary segmentation methods \citep{lime}, highlighted areas often conflict with human intuition \citep{kim2022hive, pitfalls}. 
Moreover, segmentation granularity has a dramatic impact on explanation quality \citep{superpixel_influence}: excessive segments introduce instability, causing LIME to produce mutually contradictory explanations for the same image and eroding trust in both the explanation mechanism and the model itself \citep{robustness, Garreau2021WhatDL,glime, bayeslime, slime}.

\textbf{This work.} We introduce \textbf{DSEG-LIME} (\textit{Data-driven Segmentation LIME}), an enhanced LIME framework that replaces traditional superpixel algorithms with flexible foundation-model segmenters. Users can plug in general-purpose models such as SAM (Segment Anything) \citep{kirillov2023segment} or domain-adapted variants through SAM adapters \citep{chen2024sam2adapterevaluatingadapting}, producing semantically coherent regions that mirror human concepts.
Building on these powerful segmenters, we propose a \textit{hierarchical segmentation} scheme that encodes a \textit{compositional object structure}. This lets users dial in the granularity of explanations, viewing, for example, a car as a single unit or decomposed into doors, wheels, or headlights. By enabling concept-level analysis at multiple scales, DSEG-LIME delivers clearer, more relevant feature highlights and allows independent evaluation of each subconcept.
As shown in \Cref{fig:dog}, when applied to an image of a dog playing guitar, DSEG-LIME distinctly pinpoints meaningful parts (e.g., the dog’s snout, the guitar’s body), demonstrating superior alignment with human intuition compared to existing LIME variants.

\textbf{Key contributions.} We advance explainable image analysis by:
\begin{itemize}
  \item \textbf{Introduce} DSEG‑LIME, the first adaptation of LIME to harness foundation segmentation models, enabling semantically coherent region proposals.
  \item \textbf{Develop} a hierarchical segmentation framework embedding a \textit{compositional object structure}, empowering users with fine‑grained control over explanation granularity—from whole objects to individual parts (e.g., doors, wheels, headlights).
  \item \textbf{Evaluate} DSEG‑LIME against conventional segmentation and LIME variants on multiple pretrained classifiers, employing quantitative fidelity metrics \citep{anecdotal} and a qualitative user study to assess both explanatory accuracy and human interpretability, while highlighting gaps between human intuition and model reasoning \citep{pitfalls, freiesleben2023dear}.
\end{itemize}

\section{Related work} \label{related}

\textbf{Region-based perturbation XAI techniques.}
LIME is among several techniques designed to explain black box models through image perturbation. \citet{fong2017interpretable} introduced a meta-predictor framework that identifies critical regions via saliency maps. Subsequently, \citet{fong2019understanding} developed the concept of extremal perturbations to address previous methods' limitations. Additionally, \citet{kapishnikov2019xrai} advanced an integrated-gradient, region-based attribution approach for more precise model explanations. More recently, \citet{escudero2023characterizing} highlight the constraints of perturbation-based explanations, advocating for the integration of semantic segmentation to enhance image interpretation.

\textbf{Instability of LIME.} The XAI community widely recognizes the instability in LIME's explanations, which stems from LIME's design \citep{robustness, glime, bayeslime, slime}. 
\citet{robustness} handle this issue by showing the instability of various XAI techniques when slightly modifying the instance to be explained. A direct improvement is Stabilized-LIME (SLIME) proposed by \citet{slime} based on the central limit theorem to approximate the number of perturbations needed in the data sampling approach to guarantee improved explanation stability. \citet{bayeslime} improve stability by exploiting prior knowledge and using Bayesian reasoning - BayLIME.
GLIME \citep{glime} addresses this issue by employing an improved local and unbiased data sampling strategy, resulting in explanations with higher fidelity - similar to the work by \citet{Rashid_Amparore_Ferrari_Verda_2024}.
Recent advancements include Stabilized LIME for Consistent Explanations (SLICE) \cite{bora2024slice}, which improves LIME through a novel feature selection mechanism that removes spurious superpixels and introduces an adaptive perturbation approach for generating neighborhood samples. 
Another hierarchical-based variation, DLIME \cite{dlime}, utilizes agglomerative hierarchical clustering to organize training data, focusing on tabular datasets. In contrast, DSEG-LIME extends this concept to images by leveraging the hierarchical structure of image segments.

\textbf{Segmentation influence on explanation.} The segmentation algorithm utilized to sample data around the instance $\mathbf{x}$ strongly influences its explanation. It directly affects the stability of LIME itself, as suggested by \citet{10082810}. This behavior is in line with the investigation by \citet{superpixel_influence} that examines the influence of different segmentation techniques in the medical domain, showing that the quality of the explanation depends on the underlying feature generation process.
\citet{blücher2024decoupling} explore how occlusion and sampling strategies affect model explanations when integrated with segmentation techniques for XAI, including LRP (Layer-Wise Relevance Propagation) \citep{montavon2019layer} and SHAP \citep{shap}. Their study highlights how different strategies provide unique explanations while evaluating the SAM technique in image segmentation. \citet{sun2023explain} use SAM within the SHAP framework to provide conceptually driven explanations, which we discuss in Appendix B.4 \cite{full_paper}. This work reinforces the strategy of employing deeper models to elucidate simpler ones by seamlessly integrating these approaches into widely adopted XAI frameworks.

\textbf{Segmentation hierarchy.} 
Hierarchical Segment Grouping (HSG) \cite{ke2022unsupervisedhierarchicalsemanticsegmentation} performs unsupervised hierarchical semantic segmentation by leveraging multiview cosegmentation and clustering transformers to enforce spatial and semantic consistency across images.
The work of \citet{li2022deep} aims to simulate the way humans structure segments hierarchically and introduce a framework called Hierarchical Semantic Segmentation Networks (HSSN), which approaches segmentation through a pixel-wise multi-label classification task. HIPPIE (HIerarchical oPen-vocabulary, and unIvErsal segmentation), proposed by \citet{wang2024hierarchical}, extends hierarchical segmentation by merging text and image data multimodally. It processes inputs through decoders to extract and then fuse visual and text features into enhanced representations.
While numerous hierarchical segmentation methods exist, their specialized, end‑to‑end pipelines cannot accommodate external foundation models—making them incompatible with model‑agnostic explainers like LIME. To bridge this gap, we introduce a lightweight, compositional hierarchy within LIME that supports plug‑and‑play integration of any foundation or domain‑specific segmenter.

\section{Foundations of LIME} \label{lime}

\begin{figure}[h]  
\centering
  \includegraphics[width = \linewidth]{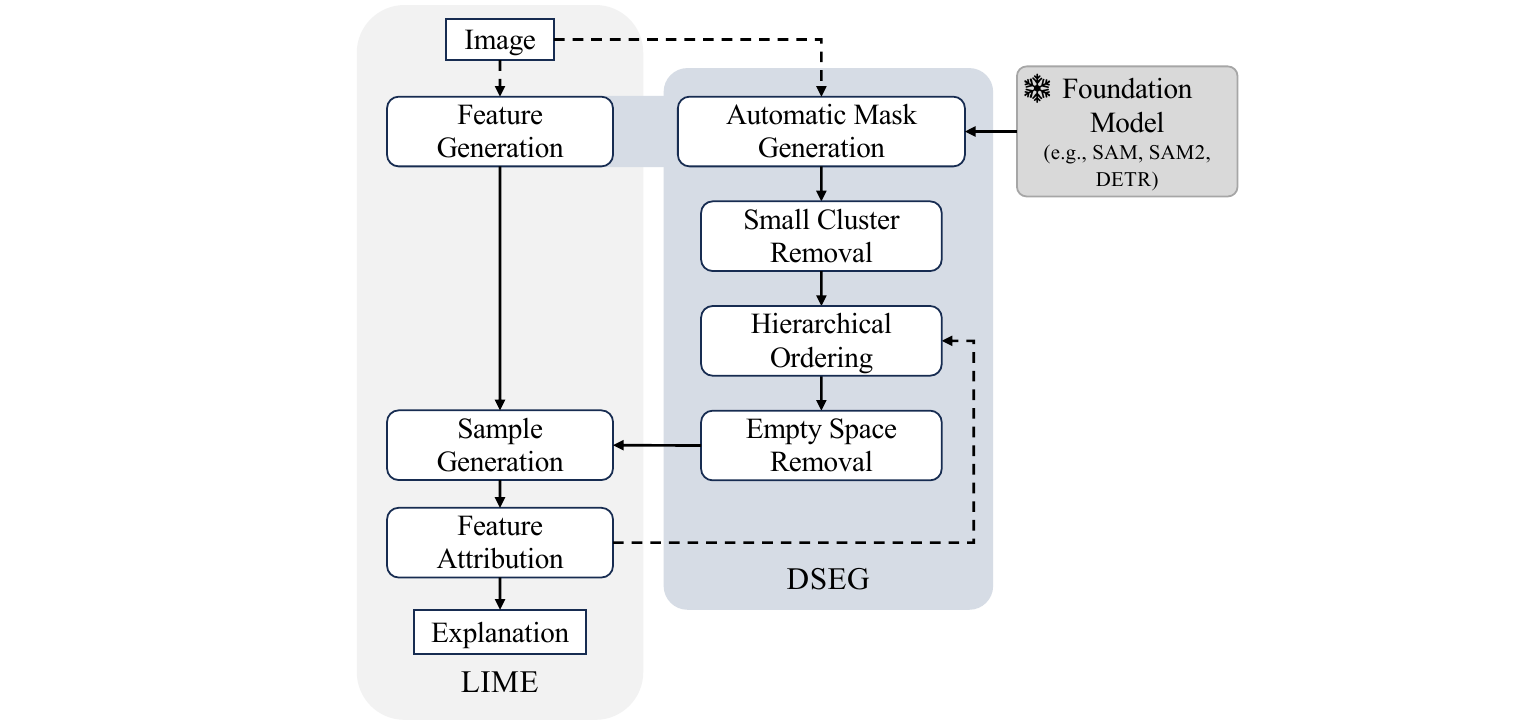}
\caption{\textbf{DSEG Pipeline in LIME.} LIME image analysis pipeline with DSEG integrated into feature generation. Dashed lines indicate the option to apply DSEG. The frozen segmentation foundation model can be freely chosen by the user.}
\label{fig:dseg}
\end{figure}

LIME is a prominent XAI framework to explain a neural network $f$ in a \textit{model-agnostic} and \textit{instance-specific} (local) manner. It applies to various modalities, such as images or text \citep{lime}. In the following, we will briefly review LIME's algorithm for treating images, also visualized in \Cref{fig:dseg} next to our adaptation with DSEG.

\textbf{Notation.} 
We consider the scenario in which we deal with image data. Let $\mathbf{x} \in \mathcal{X}$ represent an image within a set of images, and let $\mathbf{y} \in \mathcal{Y}$ denote its corresponding label in the output space with logits $\mathcal{Y} \subseteq \mathbb{R}$ indicating the labels in $\mathcal{Y}$. We denote the neural network we want to explain by $f: \mathcal{X} \rightarrow \mathcal{Y}$. This network functions by accepting an input $\mathbf{x}$ and producing an output in $\mathcal{Y}$, which signifies the probability $p$ of the instance being classified into a specific class. 

\textbf{Feature generation.}
The technique involves training a local, interpretable surrogate model $g \in G$, where $G$ is a class of interpretable models, such as linear models or decision trees, which approximates $f$'s behavior around an instance $\mathbf{x}$ \citep{lime}. This instance needs to be transformed into a set of features that can be used by $g$ to compute the importance score of its features. In the domain of imagery data, segmentation algorithms segment $\mathbf{x}$ into a set of superpixels $s_0 ...s_d \in \mathcal{S}^D$, done by conventional techniques \citep{10.5555/3454287.3454867}. We use these superpixels as the features for which we calculate their importance score. This step illustrates the motivation of \Cref{problem}, which underpins the quality of features affecting the explanatory quality of LIME.

\textbf{Sample generation.}
For sample generation, the algorithm manipulates superpixels by toggling them randomly. Specifically, each superpixel $s_i$ is assigned a binary state, indicating this feature's visibility in a perturbed sample $\mathbf{z}$. The presence (1) or absence (0) of these features is represented in a binary vector $\mathbf{z'_i}$, where the $i$-th element corresponds to the state of the $i$-th superpixel in $\mathbf{z}$.
When a feature $s_i$ is absent (i.e., $s_i = 0$), its pixel values in $\mathbf{z}$ are altered. This alteration typically involves replacing the original pixel values with a non-information holding value, such as the mean pixel value of the image or a predefined value (e.g., black pixels) \citep{lime, glime}. 

\textbf{Feature attribution.}
LIME employs a proximity measure, denoted as $\pi_\mathbf{x}$, to assess the closeness between the predicted outputs $f(\mathbf{z})$ and $f(\mathbf{x})$, which is fundamental in assigning weights to the samples. In the standard implementation of LIME, the kernel $\pi_\mathbf{x}(\mathbf{z})$ is as $\pi_\mathbf{x}(\mathbf{z'}) = \exp\left(-\frac{D(\mathbf{x'},\mathbf{z'})^2}{\sigma^2}\right)$, where $\mathbf{x'}$ is a binary vector, all states are set to 1, representing the original image $\mathbf{x}$. $D$ represents the $L2$ distance, given by $D(\mathbf{x'}, \mathbf{z'}) = \sqrt{\sum_{i=1}^{n} (\mathbf{x'}_i - \mathbf{z'}_i)^2}$ and $\sigma$ being the width of the kernel. Subsequently, LIME trains a linear model, minimizing the loss function $\mathcal{L}$, which is defined as:
\begin{equation}
\mathcal{L}(f, g, \pi_\mathbf{x}) = \sum_{\mathbf{z}, \mathbf{z'} \in \mathcal{Z}} \pi_\mathbf{x}(\mathbf{z}) \cdot (f(\mathbf{z}) - g(\mathbf{z'}))^2
\label{eq:3_}
\end{equation}
In this equation, $\mathbf{z}$, and $\mathbf{z'}$ are sampled instances of the perturbed data set $\mathcal{Z}$, and $g$ is the interpretable model being learned \citep{lime, glime}. Interpretability is derived primarily from the coefficients of $g$. These coefficients quantify the influence of each feature on the model's prediction, with each coefficient's magnitude and direction (positive or negative) indicating the feature's relative importance. 
DSEG breaks up this fixed structure and allows for a repeated calculation of features based on the concept hierarchy and user preferences.

\section{DSEG-LIME}

In this section, we will present DSEG-LIME's two contributions: first, the substitution of traditional feature generation with a data-driven segmentation
approach (\Cref{dseg}), and second, the establishment of a hierarchical structure that organizes segments in a compositional manner (\Cref{hierarchy}).

\subsection{Data-driven segmentation integration} \label{dseg}
DSEG-LIME expands the LIME feature generation phase by incorporating data-driven segmentation models, outperforming conventional graph- or cluster-based segmentation techniques in creating recognizable image segments across various domains. For the remainder of the paper, we mainly use SAM \citep{kirillov2023segment} due to its remarkable capability to segment images in diverse areas. However, in the appendix, we show that it can also be applied to other segmentation models. 
\Cref{fig:dseg} illustrates DSEG’s integration into the LIME framework of \Cref{lime}. 
The input image is first passed to a segmentation foundation model, which outputs a set of segments. 
Importantly, we only use the generated segments for the explanation process and do not incorporate any additional information from the foundation model.
By redefining the feature basis, DSEG alters the loss in \Cref{eq:3_} and the proximity metric for sampling, often producing a surrogate $g$ that more faithfully approximates the original model $f$ at instance $\mathbf{x}$. 
However, the effect of DSEG is comparable to other segmentation methods such as SLIC, since the surrogate model relies solely on the resulting segments and does not incorporate any additional information from the underlying segmentation foundation models.

\subsection{Hierarchical segmentation} \label{hierarchy}

The segmentation capabilities of foundation models like SAM, influenced by its design and hyperparameters, allow fine and coarse segmentation of an image \citep{kirillov2023segment}. These models can segment a human-recognized concept at various levels, from the entirety of a car to its components, such as doors or windshields.
This multitude of segments enables the composition of a concept into its sub-concepts, creating a hierarchical segmentation.

\begin{figure}[h]
  \centering
  \includegraphics[width = \columnwidth]{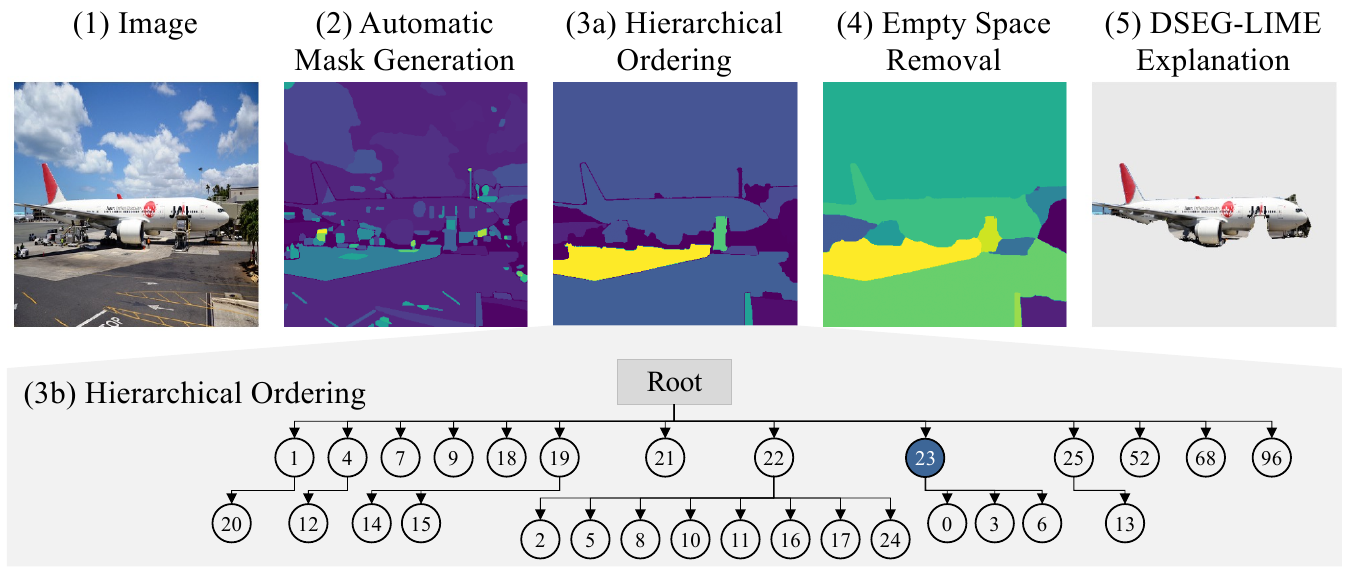}
  \caption{\textbf{Visualized DSEG pipeline.} Image (1) serves as the initial input, leading to its automatic segmentation depicted in (2). The hierarchical tree generated from this segmentation is illustrated in (3b), and (3a) showcases the mask composed of first-order nodes. Image (4) displays the finalized mask created after eliminating empty spaces, which is fed back into the sample generation of LIME. Image (5) represents the resultant explanation within the DSEG-LIME framework. The image shows an instance classified as an 'Airliner' ($p$ = 0.86) by EfficientNetB4. Node 23 (blue node) indicates the segment that represents the superpixel of the airliner. 
  }
  \label{fig:pipeline_images}
\end{figure}

Our proposed framework, DSEG, operates in successive stages. To harness this hierarchy within LIME, we introduce a depth parameter $d$ that allows users to specify the desired level of segmentation granularity for more personalized explanations. At $d=1$, we compute the importance scores for all segments (coarse) of the top level. Segments with top $k$ scores, a user‐defined value, are then subdivided into their immediate children, and at $d=2$ we recompute importance scores on these finer segments. This refine–and–rescore cycle continues, incrementing $d$ at each iteration, until the user’s maximum depth is reached or no further subdivisions exist. By allowing the surrogate model to "zoom in" iteratively on the most salient human‐recognizable parts, DSEG produces explanations that are both semantically rich and tailored to the user’s needs.
Next, we detail each step of the DSEG pipeline (\ref{fig:dseg}), illustrate its intermediate outputs (\ref{fig:pipeline_images}), and present the complete pseudocode for clarity (\cref{pseudocode}).

\textbf{Automatic mask generation.} 
Masks, also called segments or superpixels, represent distinct regions of an image. In the following, we denote the segmentation foundation model, such as SAM, by $\zeta$. Depending on the foundation model employed, $\zeta$ can be prompted using various methods, including points, area markings, text inputs, or automatically segmenting all visible elements in an image. For the main experiments of DSEG, we utilize the last prompt, automated mask generation, since we want to segment the whole image for feature generation without human intervention. We express the process as follows:
\begin{equation}
M_{\text{auto}} = \zeta(\mathbf{x}, G_{\text{prompt}}) \text{, with } \mathcal{S^D} = M_{\text{auto}},
\end{equation}
where $\mathbf{x}$ denotes the input image, and $G_{\text{prompt}}$ specifies a general prompt configuration designed to enable automated segmentation. The output, $M_{\text{auto}}$, represents the automatically generated mask, as shown in \Cref{fig:pipeline_images} (2). For this work, we used SAM with a grid overlay, parameterized by the number of points per side, to facilitate the automated segmentation process. 

\textbf{Small cluster removal.} 
The underlying foundation model generates segments of varying sizes. We define a threshold $\theta$ such that segments with pixel-size below $\theta$ are excluded:
\begin{equation}
\mathcal{S'} = \{s_i \in \mathcal{S^D} \,|\, \text{size}(s_i) \geq \theta\}.
\label{eq:44}
\end{equation}
In this study, we set $\theta = 500$ to reduce the feature set (see \Cref{ablation_study}). The remaining superpixels in $\mathcal{S'}$ are considered for feature attribution. 
We incorporate this feature into DSEG to enable user-driven segment exclusion during post-processing, giving users control over the granularity within the segmentation hierarchy. This ensures that users can tailor the segmentation to their specific needs, thereby enhancing the method's flexibility and adaptability.

\textbf{Hierarchical ordering.} 
To handle overlapping segments, we impose a tree hierarchical structure $\mathcal{T} = (\mathcal{V}, \mathcal{E})$. In this structure, the overlap signifies that the foundation model has detected a sub-segment within a larger segment, representing the relationship between fine and coarse segments. The final output of DSEG, utilized for feature calculation, excludes overlapping segments.
The nodes $v \in \mathcal{V}$ denote segments in $\mathcal{S'}$, and the edges $(u,v) \in \mathcal{E}$ encode the hierarchical relationship between segments. This hierarchical ordering process $H(\mathcal{S'})$ is a composition of the relative overlap of the segments, defined as:
\begin{equation}
H(\mathcal{S'}) = \text{BuildHierarchy}(\mathcal{S'}, \text{OverlapMetric}),
\label{eq:3}
\end{equation}
where OverlapMetric quantifies the extent of overlap between two segments $s_1, s_2 \in \mathcal{S'}$ defined by
\begin{equation}
\text{OverlapMetric}(s_1, s_2) = \frac{|s_1 \cap s_2|}{| s_2|}.
\label{eq:4}
\end{equation}
The hierarchy prioritizes parent segments (e.g., person) over child segments (e.g., clothing), as depicted (3a) in \Cref{fig:pipeline_images}. Each node represents one superpixel with its unique identifier. \Cref{hcompo} presents the BuildHierarchy method in greater detail. 
The depth $d$ of the hierarchy determines the granularity of the explanation, as defined by the user. A new set $\mathcal{S}'_d$, with $d = 1$, includes all nodes below the root. For $d > 1$, DSEG does not start from the beginning. Instead, it uses the segmentation hierarchy and segments $\mathcal{S'}$ from the first iteration. It adds the child nodes of the top-$k$ most significant parent nodes in $\mathcal{S}’_d$ at depth $d - 1$, identified during feature attribution. These are the only nodes used to train the surrogate model.
We visualize this selection in the tree shown in \Cref{fig:pipeline_images} (3b), where all nodes with depth one, including the children of node 23, are considered in the second iteration. 
For the scope of this paper, we concentrate on the first-order hierarchy ($d = 1$) but provide additional explanations with $d = 2$ in \Cref{depth2}. 

\textbf{Empty space removal.} 
In hierarchical segmentation, some regions occasionally remain unsegmented. We refer to these areas as $R_{\text{unseg}}$. To address this, we employ the nearest neighbor algorithm, which assigns each unsegmented region in $R_{\text{unseg}}$ to the closest segment within the set $\mathcal{S}'_d$:
\begin{equation}
\mathcal{S}_d = \text{NearestNeighbor}(R_{\text{unseg}}, \mathcal{S}'_d).  
\label{33}
\end{equation}
Although this modifies the distinctiveness of concepts, it enhances DSEG-LIME's explanatory power. 
DSEG then utilizes the features $s_0, \ldots, s_d \in \mathcal{S}_d$ for feature attribution within LIME. \Cref{fig:pipeline_images} (4) shows the corresponding mask along with the explanation of $d = 1$ in step (5) for the 'airliner' class. An ablation study of these steps is in \Cref{ablation_study} and in \Cref{feature_maps} we show exemplary feature attribution maps.

\section{Evaluation}
In the following section, we will outline our experimental setup (\Cref{segmentations}) and introduce the evaluation framework designed to assess DSEG-LIME both quantitatively (\Cref{eval_section}) and qualitatively (\Cref{user}), compared to other LIME methodologies utilizing various segmentation algorithms. Subsequently, we discuss the limitations of DSEG (\Cref{limits})

\subsection{Experimental setup} \label{setup}
\textbf{Segmentation algorithms.}\label{segmentations}
Our experiment encompasses, along with SAM (vit\_h), four conventional segmentation techniques: \textit{Simple Linear Iterative Clustering} (SLIC) \citep{6205760}, \textit{Quickshift} (QS) \citep{10.5555/3454287.3454867}, \textit{Felzenszwalb} (FS) \citep{Felzenszwalb2004} and \textit{Watershed} (WS) \citep{6976891}. We carefully calibrate the hyperparameters (e.g., kernel size, marker placement, number of segments, minimum segment size) to produce segment counts comparable to those generated by SAM. This calibration ensures that no technique is unfairly advantaged due to a specific segment count, for instance, scenarios where fewer but larger segments might yield better explanations than many smaller ones. 
In the supplementary material, we demonstrate the universal property of integrating other segmentation foundation models within DSEG by presenting additional experiments with DETR \citep{DETR} and SAM 2 \citep{ravi2024sam2segmentimages} in \Cref{detr}, \Cref{sam_2}. 

\textbf{Models to explain.} The models investigated in this paper rely on pre-trained models, as our primary emphasis is on explainability. We chose EfficientNetB4 and EfficientNetB3 \citep{tan2019efficientnet} as the ones treated in this paper, where we explain EfficientNetB4 and use EfficientNetB3 for a contrastivity check \citep{anecdotal} (\Cref{quanti}). To verify that our approach works on arbitrary pre-trained models, we also evaluated it using ResNet-101 \citep{DBLP:journals/corr/HeZRS15, torchvision2016} (\Cref{res}), VisionTransformer (ViT-384) \citep{DBLP:journals/corr/abs-2010-11929} (\Cref{vis_}), and ConvNext (Tiny-224) (\Cref{ConvNext})\cite{DBLP:journals/corr/abs-2201-03545}. Furthermore, we demonstrate the applicability of our approach on a zero-shot learning example of CLIP \citep{radford2021learningtransferablevisualmodels} using a new dataset with other classes (\Cref{clip_}). 

\textbf{Dataset.}\label{dataset} We use images from the ImageNet classes \citep{5206848}, on which the covered models were trained \citep{DBLP:journals/corr/abs-2010-11929, DBLP:journals/corr/HeZRS15, tan2019efficientnet}. 
Our final dataset consists of 100 carefully selected instances (\Cref{app_data}), specifically chosen to comprehensively evaluate the techniques quantitatively. We aligned with this number to maintain consistency with the sample size used in previous studies \cite{sun2023explain, glime}. However, we want to emphasize that the selection of images is not biased toward any model. We also test the approach for another dataset in \Cref{clip_}.

\begin{table*}[ht!]

  \caption{\textbf{Quantitative summary - classes.} The table presents four quantitative areas and their metrics, comparing five segmentation techniques applied to EfficientNetB4: DSEG with SAM and comparative methods SLIC, Quickshift (QS), Felzenszwalb's (FS), and Watershed (WS). We test each with four LIME framework variations: LIME (L), SLIME (S), GLIME (G), and BayLIME (B). The experimental setup and metrics are detailed in \Cref{setup} and \Cref{quanti}. The table shows class metrics, each with a max score of 100. Arrows indicate performance improvement, highlighting the best scores for each metric bold.}
\centering
\large
\label{tab:eval}
\resizebox{0.9\linewidth}{!}{%
\begin{tabular}{ll|rrrr|rrrr|rrrr|rrrr|rrrr}
\toprule
\multirow{2}{*}{Domain} &
  \multirow{2}{*}{Metric} &
  \multicolumn{4}{c}{DSEG} &
  \multicolumn{4}{c}{SLIC} &
  \multicolumn{4}{c}{QS} &
  \multicolumn{4}{c}{FS} &
  \multicolumn{4}{c}{WS} \\ \cmidrule(r){3-6} \cmidrule(r){7-10} \cmidrule(r){11-14} \cmidrule(r){15-18} \cmidrule(r){19-22} 
                                     &                       & L & S & G & B & L & S & G & B & L & S & G & B & L & S & G & B & L & S & G & B \\ 
\midrule
\multirow{3}{*}{Correctness}         & Random Model $\uparrow$
         &  \textbf{74}   & \textbf{74} & \textbf{74} & \textbf{74} & 68 & 67 & 68 & 68 & 70 & 70 & 70 & 70& 69 & 70 & 70 & 70 & 67 & 66 & 67 & 66 \\

                                     & Random Expl. $\uparrow$
         & 86   &90 & \textbf{93} & 88 & 81 &  79 & 80 & 84 & 76 & 82 & 75 & 81 & 87 & 83 & 85 & 81 & 79 & 81 & 86 & 85 \\

                                     & Single Deletion $\uparrow$
         & 63   & 62 & \textbf{64} & 63 & 36 & 34 & 33 & 34 & 19 & 20 & 17 & 18 & 29 & 30 & 27 & 30 & 24 & 27 &26 & 24 \\
\midrule
\multirow{2}{*}{\begin{tabular}[c]{@{}l@{}}Output \\ Completeness\end{tabular}} & Preservation $\uparrow$
         & 77 & 74 & 73 & 73 & 74 & 74 & 75 & 74 & 68 & 65 & 67 & 64 & 71 & 74 & 74 & 72 & \textbf{79} & \textbf{79} & 77 & 78 \\

                                     & Deletion $\uparrow$
         & 72 & \textbf{74} & \textbf{74} & \textbf{74} & 39 & 39 & 39 & 40 & 33 & 34 & 35 & 34 & 43 & 43 & 43 & 43 & 44 & 44 & 44 & 44 \\
\midrule
\multirow{1}{*}{Consistency}         & Noise Preservation $\uparrow$
         & 76 & 74 & \textbf{77} & \textbf{77} & 72 & 72 & 72 & 72 & 58 & 57 & 60 & 57 & 62 & 64 & 63 & 63 & 71 & 70 & 71 & 71 \\
         & Noise Deletion $\uparrow$
         & \textbf{68} & 66 & 66 & 66 & 40 & 41 & 40 & 40 & 32 & 34 & 33 & 34 & 39 & 41 & 39 & 41 & 48 & 49 & 48 & 49 \\
\midrule

\multirow{2}{*}{Contrastivity}       & Preservation $\uparrow$
         & 64 & 63 & 64&\textbf{65} & 54 & 54 & 55 & 55 & 49 & 46 & 46 & 45 & 49 & 52 & 53 & 52 & 54 & 54 & 54 & 53 \\

                                     & Deletion $\uparrow$
         & 69 & 70 & \textbf{71} & \textbf{71} & 49 & 50 & 48 & 50 & 39 & 39 & 40 & 41 & 42 & 42 & 42 & 42 & 47 &47 & 47 & 47 \\
\bottomrule
\end{tabular}
}%
\end{table*}

\textbf{Hyperparameters and hardware setup.}\label{hyper} The experiments were conducted on an Nvidia RTX A6000 GPU. 
We compare standard LIME, SLIME \citep{slime}, GLIME \citep{glime}, and BayLIME \citep{bayeslime}, all integrated with DSEG, using 256 samples per instance, a batch size of ten and mean superpixel value for perturbation. For each explanation, up to three features are selected based on their significance, identified by values that exceed the average by more than 1.5 times the standard deviation. In BayLIME, we use the ‘non-info-prior’ setting to avoid introducing priors benefiting one method, even though the original papers show that this can have a diminishing effect.
For SAM, we configure it to use 32 points per side, and conventional segmentation techniques are adjusted to achieve a similar segment count, as previously mentioned. 
In SLIC, we modify the number of segments and compactness; in Quickshift, the kernel size and maximum distance; in Felzenszwalb, the scale of the minimum size parameter; and in Watershed, the number of markers and compactness.
Other hyperparameters remain at default settings to ensure a balanced evaluation across methods. Additional comparisons with SLICE \cite{bora2024slice} and EAC \cite{sun2023explain} are in the supplementary material.

\subsection{Quantitative evaluation} \label{eval_section}

We adapt the framework by \citet{anecdotal} to quantitatively assess XAI outcomes in this study, covering three domains: \textit{content}, \textit{presentation}, and \textit{user experience}. In the content domain, we evaluate \textit{correctness}, \textit{output completeness}, \textit{consistency}, and \textit{contrastivity}. Presentation domain metrics like \textit{compactness} and \textit{confidence} are assessed under content for simplicity. We will briefly describe each metric individually to interpret the results correctly. The user domain, detailed in \Cref{user}, includes a user study that compares our approach with other segmentation techniques in LIME. We use quantitative and qualitative assessments to avoid over-emphasizing technical precision or intuitive clarity \citep{pitfalls}.

\subsubsection{Quantitative metrics definition} \label{quanti}

\textbf{Correctness} involves two randomization checks. The \textit{model randomization parameter check} (Random Model) \citep{10.5555/3495724.3495784} tests if changing the random model parameters leads to different explanations. The \textit{explanation randomization check} (Random Expl.) \citep{10.5555/3495724.3497370} examines if random output variations in the predictive model yield various explanations. For both metrics, in \Cref{tab:eval} we count the instances where explanations result in different predictions when reintroduced into the model under analysis. The domain also utilizes two deletion techniques: \textit{single deletion} \citep{ijcai2020p63} and \textit{incremental deletion} \citep{pmlr-v97-goyal19a, 10.5555/3454287.3454867}. \textit{Single deletion} serves as an alternative metric to assess the completeness of the explanation, replacing less relevant superpixels with a specific background to evaluate their impact on the model predictions \citep{10.5555/3495724.3496225}. After these adjustments, we note instances where the model maintains the correct image classification.
\textit{Incremental deletion} (Incr. Deletion) entails progressively eliminating features from most to least significant based on their explanatory importance. We observe the model's output variations, quantifying the impact by measuring the area under the curve (AUC) of the model's confidence, as parts of the explanation are excluded. This continues until a classification change is observed (not ground truth class), and the mean AUC score for this metric is documented in \Cref{tab:example}.

\textbf{Output completeness}\label{oc} measures whether an explanation covers the crucial area for accurate classification. It includes a \textit{preservation check} (Preservation) \citep{pmlr-v97-goyal19a} to assess whether the explanation alone upholds the original decision, and a \textit{deletion check} (Deletion) \citep{zhang2023xai} to evaluate the effect of excluding the explanation on the prediction outcome \citep{10.5555/3495724.3496225}. This approach assesses both the completeness of the explanation and its impact on the classification. The results are checked to ensure that the consistency of the classification is maintained. \textit{Compactness} is also considered, highlighting that the explanation should be concise and cover all the areas necessary for prediction \citep{chang2018explaining}, reported by the mean value.

\textbf{Consistency} assesses explanation robustness to minor input alterations, like Gaussian noise addition, by comparing pre-and post-perturbation explanations for \textit{stability against slight changes} (Noise Stability) \citep{9472817}, using both preservation and deletion checks. For consistency of the feature importance score, we generate explanations for the same instance 16 times (Rep. Stability), calculate the standard deviation $\sigma_{i}$ for each coefficient $i$, and then average all $\sigma_{i}$ values. This yields $\bar{\sigma}$, the average standard deviation of coefficients, and is reported as the mean score. Furthermore, we reference \cite{bora2024slice} and compute a Gini coefficient to assess the inequality of the coefficients of the superpixels. We also evaluate DSEG directly within SLICE in \Cref{slice_chapt}.

\textbf{Contrastivity} integrates several previously discussed metrics, aiming for \textit{target-discriminative} explanations. This means that an explanation $e_\textbf{x}$ for an instance $\textbf{x}$ from a primary model $f_1$ (EfficientNetB4) should allow a secondary model $f_2$ (EfficientNetB3) to mimic the output of $f_1$ as $f_1(\textbf{x}) \approx f_2(e_\textbf{x})$ \citep{10.5555/3454287.3455204}. The approach checks the explanation's utility and transferability across models, using EfficientNetB3 for preservation and deletion tests to assess consistency.

\begin{table*}[ht!]
\caption{\textbf{Quantitative summary - numbers.} The table summarizes metrics from \Cref{quanti}, focusing on those quantified by rational numbers like incremental deletion, compactness, representational stability, and average computation time across the examples, detailed in \Cref{setup}. Once more, the arrows denote the direction of improvement performance.}
\centering
\label{tab:example}
\resizebox{0.9\linewidth}{!}{%
\Large
\begin{tabular}{l|rrrr|rrrr|rrrr|rrrr|rrrr}
\toprule
\multirow{2}{*}{Metric} & \multicolumn{4}{c}{DSEG}   & \multicolumn{4}{c}{SLIC}  & \multicolumn{4}{c}{QS}    & \multicolumn{4}{c}{FS}    & \multicolumn{4}{c}{WS}    \\
\cmidrule(r){2-5} \cmidrule(r){6-9} \cmidrule(r){10-13} \cmidrule(r){14-17} \cmidrule(r){18-21}
                        & L  & S    & G    & B    & L    & S    & G    & B    & L    & S    & G    & B    & L    & S    & G    & B    & L    & S    & G    & B    \\
                        \midrule
Gini $\uparrow$ & 0.53 & 0.52 & \textbf{0.54} & 0.53 & 0.49 & 0.49 & 0.50 & 0.50 & 0.41 & 0.42 & 0.41 & 0.41 & 0.45 & 0.44 &0.45 & 0.45 & 0.46 & 0.47 & 0.48 & 0.47 \\

Incr. Deletion $\downarrow$         & 1.12 & \textbf{0.44} & 0.45 & 0.50 & 0.81 & 0.82 & 0.81 & 0.82 & 1.57 & 1.58 & 1.60 & 1.61 & 1.49 & 1.48 & 1.54 & 1.52 & 0.79 & 0.80 & 0.79 & 0.78 \\

Compactness $\downarrow$            & 0.17 & 0.18 & 0.18 & 0.18 & 0.15 & 0.15 & 0.15 & 0.15 & \textbf{0.12} & \textbf{0.12} & \textbf{0.12} & \textbf{0.12} & \textbf{0.12} & \textbf{0.12} & 0.13 & 0.13 & \textbf{0.12} & \textbf{0.12} & \textbf{0.12} & 0.13 \\

Rep. Stability $\downarrow$         & \textbf{.010} & \textbf{.010} & \textbf{.010} & \textbf{.010} & .011 & .011 & .012 & .012 & .011 & .011 & .012 & .011 & .012 & .011 & .012 & .011 & .012 & .012 & .012 & .012 \\

Time $\downarrow$                   & 28.5 & 31.1 & 31.9 & 31.7 & 16.1 & 16.2 & 16.8 &16.8 & 28.6 & 28.6 & 28.9 & 28.6 & 15.9 & 16.0 & 17.3 & 16.7 & 15.7 & \textbf{15.6} & 17.0 & 16.5 \\
\bottomrule

\end{tabular}%
}
\end{table*}

\subsubsection{Quantitative evaluation results} 
\Cref{tab:eval} presents the outcomes of all metrics associated with class-discriminative outputs. The numbers in bold signify the top results, with an optimal score of 100. 
We compare LIME (L) \citep{lime} with the LIME techniques discussed in \Cref{related}, SLIME (S) \citep{slime}, GLIME (G) \citep{glime}, and BayLIME (B) \citep{bayeslime} in combination with DSEG and the segmentation techniques from \Cref{segmentations}.
The randomization checks in the correctness category confirm that the segmentation algorithm bias does not inherently affect any model. This is supported by the observation that most methods correctly misclassify when noise is introduced or the model's weights or predictions are shuffled.
In contrast, DSEG excels in other metrics, surpassing alternative methods regardless of the LIME technique applied. In the output completeness domain, DSEG's explanations more effectively capture the critical areas necessary for the model to accurately classify an instance, whether by isolating or excluding the explanation. This efficacy is supported by the single deletion metric, akin to the preservation check but with a perturbed background. Moreover, noise does not compromise the consistency of DSEG's explanations. The contrastivity metric demonstrates DSEG's effectiveness in creating explanations that allow another AI model to produce similar outputs in over half of the cases and outperform alternative segmentation approaches. 
Overall, the impact of the LIME feature attribution calculation remains relatively consistent, as we possess an average of 21.07 segments in the dataset under evaluation.

\Cref{tab:example} further illustrates DSEG's effectiveness in identifying key regions for model output, especially in scenarios of incremental deletion where GLIME outperforms. Bolded values represent the best performance in their according category. Although compactness metrics show nearly uniform segment sizes across the techniques, smaller segments do not translate to better performance in other areas. Repeated experimentation suggests that stability is less influenced by the LIME variant (with the chosen hyperparameters) and more by the segmentation approach. 
In addition, the Gini value highlights that the feature values are more distinct compared to others.
Further experiments show that DSEG outperforms the other techniques regarding stability as the number of features increases. This advantage arises from the tendency of data-driven approaches to represent known objects uniformly as a single superpixel (\Cref{app_stab}). Thus, if a superpixel accurately reflects the instance that the model in question predicts, it can be accurately and effortlessly matched with one or a few superpixels - such accurate matching leads to a more precise and more reliable explanation. Conventional segmentation algorithms often divide the same area into multiple superpixels, creating unclear boundaries and confusing differentiation between objects. 
DSEG has an average runtime of approximately 30 seconds across different settings. Although the standard LIME approach using SLIC operates at around 16 seconds, the additional computational steps in DSEG (e.g., SAM integration and hierarchical ordering) introduce only a modest overhead. Importantly, when compared to the widely used Quickshift option, which typically takes about 29 seconds, DSEG’s runtime is directly comparable. Thus, the extra computational cost is minimal relative to the substantial benefits in explanation quality.

The supplementary experiments (see \cref{model_evals}) exhibit similarly consistent quantitative gains. We also extend our analysis by benchmarking against a SHAP‑based explainer \citep{sun2023explain}, integrating DSEG into the SLICE framework \citep{bora2024slice}, and testing different segmentation models. 

\subsection{Qualitative evaluation} \label{user}

\textbf{User study.} Following the methodology by \citet{chromik2020taxonomy}, we conducted a user study (approved by the institute's ethics council) to assess the interpretability of the explanations. This study involved 87 participants recruited via Amazon Mechanical Turk (MTurk) and included 20 random images in our dataset (\Cref{app_data}). These images were accompanied by explanations using DSEG and other segmentation techniques within the LIME framework (\Cref{hyper}).

\begin{table}[h] 
\caption{\textbf{User study results.} This table summarizes each segmentation approach's average scores and top-rated counts of the user study results.}
\label{user_table}
\centering
\resizebox{0.9\columnwidth}{!}{
\scriptsize 
\begin{tabular}{l|rrrrr}
\toprule
Metric & DSEG & SLIC & QS & FS & WS \\
\midrule
Avg. Score $\uparrow$ & \textbf{4.16} & 3.01 & 1.99 & 3.25 & 2.59 \\
Best Rated $\uparrow$ & \textbf{1042} & 150 & 90 & 253 & 205 \\
\bottomrule
\end{tabular}
}
\end{table}

Participants rated the explanations on a scale from 1 (least effective) to 5 (most effective) based on their intuitive understanding and the predicted class. \Cref{user_table} summarizes the average scores, the cumulative number of top-rated explanations per instance, and the statistical significance of user study results for each segmentation approach. DSEG is most frequently rated as the best and consistently ranks high, even when it is not the leading explanation. Paired t-tests indicate that DSEG is statistically significantly superior (additional results in \Cref{user_appendix}).

\begin{figure}[h]
  \centering
  \includegraphics[width=0.99\linewidth]{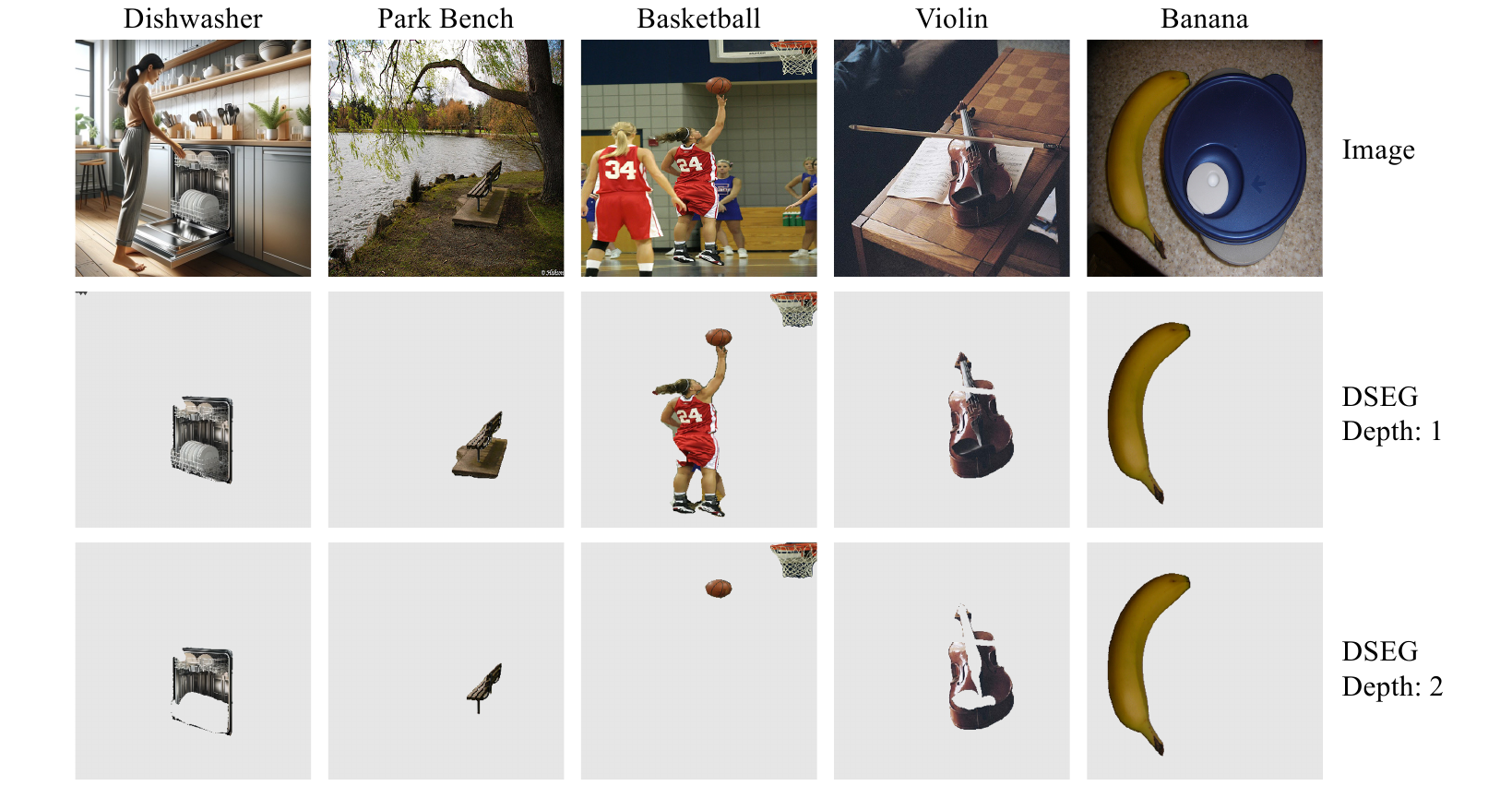}
  \caption{\textbf{DSEG examples.} Exemplary images from the evaluation dataset, illustrating DSEG explanations at $d = 1$, and $d = 2$.}
  \label{fig:d2}
\end{figure}

\textbf{Exemplary Explanations.}
\Cref{fig:d2} displays five examples from our evaluation, with the top images showing DSEG's explanations at a hierarchy depth of one and the bottom row at a depth of two. These explanations demonstrate that deeper hierarchies focus on smaller regions. However, the banana example illustrates a scenario where no further segmentation occurs if the concept, like a banana, lacks sub-components for feature generation, resulting in identical explanations at both depths. An example with a depth of three applied to a medical CT image is presented in \Cref{med_3}.

These five images (with $d = 1$) were also included in the user study. For the complete set of 100 evaluation images and their attribution maps, please refer to the supplementary material (\Cref{feature_maps}).

\subsection{Limitations and future work}\label{limits} DSEG-LIME performs feature generation directly on images before inputting them into the model for explanation. For models like ResNet with smaller input sizes \citep{DBLP:journals/corr/HeZRS15}, the quantitative advantages are less evident (\Cref{res}). Experiments have shown that better results can be achieved with a lower SAM stability score threshold.
Furthermore, substituting superpixels with a specific value in preservation and deletion evaluations can introduce an inductive bias \citep{Garreau2021WhatDL}. To reduce this bias, using a generative model to synthesize replacement areas could offer a more neutral alteration.
Additionally, future work should thoroughly evaluate feature attribution maps to ensure that methods assign significant attributions to the correct regions, as in \cref{feature_maps}. This comprehensive assessment is essential to verify the interpretability and reliability of such methods.
Lastly, our approach, like any other LIME-based method \citep{lime, glime, bayeslime, slime}, does not assume a perfect match between the explanation domains and the model’s actual domains since it simplifies the model by a local surrogate. Nonetheless, our quantitative analysis confirms that the approximations closely reflect the model's behavior. Future work could focus on integrating foundation models directly into the system through a model-intrinsic approach, similar to \citep{sun2023explain}.

\textbf{No free lunch.} While DSEG shows promising results across various domains, it is not universally applicable—its performance depends on the underlying segmentation model. If the model fails, for instance, due to a lack of domain-specific knowledge or the complexity of the feature generation task, the resulting concept extraction can be suboptimal (\Cref{fail_dseg}). To address this limitation, future work could explore alternative segmentation backbones, such as HSSN \citep{wang2024hierarchical} or HIPPIE \citep{li2022deep}, in place of SAM (or DETR). Moreover, recent extensions of SAM have introduced domain-adaptive modules that can improve segmentation in specialized contexts \citep{chen2024sam2adapterevaluatingadapting}.

\section{Conclusion}

In this study, we introduced DSEG-LIME, an extension to the LIME framework, incorporating segmentation foundation models for feature generation with hierarchical feature calculation.  This approach ensures that the generated features more accurately reflect human-recognizable concepts, enhancing the interpretability of explanations. Furthermore, we refined the process of feature attribution within LIME through an iterative method, establishing a segmentation hierarchy that contains the relationships between components and their subcomponents. In \Cref{wrong_class}, we show that our idea also helps explain a model's wrong classifications. Through a comprehensive two-part evaluation, split into quantitative and qualitative analyses, DSEG emerged as the superior method, outperforming other LIME-based approaches in most evaluated metrics.
The adoption of foundational models marks a significant step toward enhancing post-hoc and model-agnostic interpretability of deep learning models.

\begin{ack}This research was supported in part by the German Federal Ministry for Economic Affairs and Climate Action of Germany (BMWK), and in part by the German Federal Ministry for Research, Technology, and Space (BMFTR).
\end{ack}


\bibliography{main}

\begin{thebibliography}{61}
\providecommand{\natexlab}[1]{#1}
\providecommand{\url}[1]{\texttt{#1}}
\expandafter\ifx\csname urlstyle\endcsname\relax
  \providecommand{\doi}[1]{doi: #1}\else
  \providecommand{\doi}{doi: \begingroup \urlstyle{rm}\Url}\fi

\bibitem[Achanta et~al.(2012)Achanta, Shaji, Smith, Lucchi, Fua, and Süsstrunk]{6205760}
R.~Achanta, A.~Shaji, K.~Smith, A.~Lucchi, P.~Fua, and S.~Süsstrunk.
\newblock Slic superpixels compared to state-of-the-art superpixel methods.
\newblock \emph{IEEE Transactions on Pattern Analysis and Machine Intelligence}, 34\penalty0 (11):\penalty0 2274--2282, 2012.
\newblock \doi{10.1109/TPAMI.2012.120}.

\bibitem[Adebayo et~al.(2020)Adebayo, Muelly, Liccardi, and Kim]{10.5555/3495724.3495784}
J.~Adebayo, M.~Muelly, I.~Liccardi, and B.~Kim.
\newblock Debugging tests for model explanations.
\newblock In \emph{Proceedings of the 34th International Conference on Neural Information Processing Systems}, NIPS'20, Red Hook, NY, USA, 2020. Curran Associates Inc.
\newblock ISBN 9781713829546.

\bibitem[Albini et~al.(2020)Albini, Rago, Baroni, and Toni]{ijcai2020p63}
E.~Albini, A.~Rago, P.~Baroni, and F.~Toni.
\newblock Relation-based counterfactual explanations for bayesian network classifiers.
\newblock In C.~Bessiere, editor, \emph{Proceedings of the Twenty-Ninth International Joint Conference on Artificial Intelligence, {IJCAI-20}}, pages 451--457. International Joint Conferences on Artificial Intelligence Organization, 7 2020.
\newblock \doi{10.24963/ijcai.2020/63}.
\newblock Main track.

\bibitem[Alvarez{-}Melis and Jaakkola(2018)]{robustness}
D.~Alvarez{-}Melis and T.~S. Jaakkola.
\newblock On the robustness of interpretability methods.
\newblock \emph{CoRR}, abs/1806.08049, 2018.

\bibitem[{Barredo Arrieta} et~al.(2020){Barredo Arrieta}, Díaz-Rodríguez, {Del Ser}, Bennetot, Tabik, Barbado, Garcia, Gil-Lopez, Molina, Benjamins, Chatila, and Herrera]{XAI_concepts}
A.~{Barredo Arrieta}, N.~Díaz-Rodríguez, J.~{Del Ser}, A.~Bennetot, S.~Tabik, A.~Barbado, S.~Garcia, S.~Gil-Lopez, D.~Molina, R.~Benjamins, R.~Chatila, and F.~Herrera.
\newblock Explainable artificial intelligence (xai): Concepts, taxonomies, opportunities and challenges toward responsible ai.
\newblock \emph{Information Fusion}, 58:\penalty0 82--115, 2020.
\newblock ISSN 1566-2535.
\newblock \doi{https://doi.org/10.1016/j.inffus.2019.12.012}.

\bibitem[Bluecher et~al.(2024)Bluecher, Vielhaben, and Strodthoff]{blücher2024decoupling}
S.~Bluecher, J.~Vielhaben, and N.~Strodthoff.
\newblock Decoupling pixel flipping and occlusion strategy for consistent {XAI} benchmarks, 2024.
\newblock ISSN 2835-8856.
\newblock URL \url{https://openreview.net/forum?id=bIiLXdtUVM}.

\bibitem[Bora et~al.(2024)Bora, Terh{\"o}rst, Veldhuis, Ramachandra, and Raja]{bora2024slice}
R.~P. Bora, P.~Terh{\"o}rst, R.~Veldhuis, R.~Ramachandra, and K.~Raja.
\newblock Slice: Stabilized lime for consistent explanations for image classification.
\newblock In \emph{Proceedings of the IEEE/CVF Conference on Computer Vision and Pattern Recognition}, pages 10988--10996, 2024.

\bibitem[Carion et~al.(2020)Carion, Massa, Synnaeve, Usunier, Kirillov, and Zagoruyko]{DETR}
N.~Carion, F.~Massa, G.~Synnaeve, N.~Usunier, A.~Kirillov, and S.~Zagoruyko.
\newblock End-to-end object detection with transformers.
\newblock In A.~Vedaldi, H.~Bischof, T.~Brox, and J.-M. Frahm, editors, \emph{Computer Vision -- ECCV 2020}, pages 213--229, Cham, 2020. Springer International Publishing.
\newblock ISBN 978-3-030-58452-8.

\bibitem[Chang et~al.(2019)Chang, Creager, Goldenberg, and Duvenaud]{chang2018explaining}
C.-H. Chang, E.~Creager, A.~Goldenberg, and D.~Duvenaud.
\newblock Explaining image classifiers by counterfactual generation.
\newblock In \emph{International Conference on Learning Representations}, 2019.

\bibitem[Chen et~al.(2023)Chen, Zhu, Deng, Cao, Wang, Zhang, Li, Sun, Zang, and Mao]{chen2023sam}
T.~Chen, L.~Zhu, C.~Deng, R.~Cao, Y.~Wang, S.~Zhang, Z.~Li, L.~Sun, Y.~Zang, and P.~Mao.
\newblock Sam-adapter: Adapting segment anything in underperformed scenes.
\newblock In \emph{Proceedings of the IEEE/CVF International Conference on Computer Vision}, pages 3367--3375, 2023.

\bibitem[Chen et~al.(2024)Chen, Lu, Zhu, Ding, Yu, Ji, Li, Sun, Mao, and Zang]{chen2024sam2adapterevaluatingadapting}
T.~Chen, A.~Lu, L.~Zhu, C.~Ding, C.~Yu, D.~Ji, Z.~Li, L.~Sun, P.~Mao, and Y.~Zang.
\newblock Sam2-adapter: Evaluating \& adapting segment anything 2 in downstream tasks: Camouflage, shadow, medical image segmentation, and more, 2024.
\newblock URL \url{https://arxiv.org/abs/2408.04579}.

\bibitem[Chromik and Schuessler(2020)]{chromik2020taxonomy}
M.~Chromik and M.~Schuessler.
\newblock A taxonomy for human subject evaluation of black-box explanations in xai.
\newblock \emph{Exss-atec@ iui}, 1, 2020.

\bibitem[Deng et~al.(2009)Deng, Dong, Socher, Li, Li, and Fei-Fei]{5206848}
J.~Deng, W.~Dong, R.~Socher, L.-J. Li, K.~Li, and L.~Fei-Fei.
\newblock Imagenet: A large-scale hierarchical image database.
\newblock In \emph{2009 IEEE Conference on Computer Vision and Pattern Recognition}, pages 248--255, 2009.
\newblock \doi{10.1109/CVPR.2009.5206848}.

\bibitem[Dosovitskiy et~al.(2021)Dosovitskiy, Beyer, Kolesnikov, Weissenborn, Zhai, Unterthiner, Dehghani, Minderer, Heigold, Gelly, Uszkoreit, and Houlsby]{DBLP:journals/corr/abs-2010-11929}
A.~Dosovitskiy, L.~Beyer, A.~Kolesnikov, D.~Weissenborn, X.~Zhai, T.~Unterthiner, M.~Dehghani, M.~Minderer, G.~Heigold, S.~Gelly, J.~Uszkoreit, and N.~Houlsby.
\newblock An image is worth 16x16 words: Transformers for image recognition at scale.
\newblock In \emph{International Conference on Learning Representations}, 2021.
\newblock URL \url{https://openreview.net/forum?id=YicbFdNTTy}.

\bibitem[Escudero-Vi{\~n}olo et~al.(2023)Escudero-Vi{\~n}olo, Besc{\'o}s, L{\'o}pez-Cifuentes, and Gaji{\'c}]{escudero2023characterizing}
M.~Escudero-Vi{\~n}olo, J.~Besc{\'o}s, A.~L{\'o}pez-Cifuentes, and A.~Gaji{\'c}.
\newblock Characterizing a scene recognition model by identifying the effect of input features via semantic-wise attribution.
\newblock In \emph{Explainable Deep Learning AI}, pages 55--77. Elsevier, 2023.

\bibitem[Felzenszwalb and Huttenlocher(2004)]{Felzenszwalb2004}
P.~F. Felzenszwalb and D.~P. Huttenlocher.
\newblock Efficient graph-based image segmentation.
\newblock \emph{International Journal of Computer Vision}, 59:\penalty0 167--181, 2004.

\bibitem[Fong et~al.(2019)Fong, Patrick, and Vedaldi]{fong2019understanding}
R.~Fong, M.~Patrick, and A.~Vedaldi.
\newblock Understanding deep networks via extremal perturbations and smooth masks.
\newblock In \emph{Proceedings of the IEEE/CVF international conference on computer vision}, pages 2950--2958, 2019.

\bibitem[Fong and Vedaldi(2017)]{fong2017interpretable}
R.~C. Fong and A.~Vedaldi.
\newblock Interpretable explanations of black boxes by meaningful perturbation.
\newblock In \emph{Proceedings of the IEEE international conference on computer vision}, pages 3429--3437, 2017.

\bibitem[Freiesleben and K{\"o}nig(2023)]{freiesleben2023dear}
T.~Freiesleben and G.~K{\"o}nig.
\newblock Dear xai community, we need to talk!
\newblock In L.~Longo, editor, \emph{Explainable Artificial Intelligence}, pages 48--65, Cham, 2023. Springer Nature Switzerland.
\newblock ISBN 978-3-031-44064-9.

\bibitem[Garreau and Mardaoui(2021)]{Garreau2021WhatDL}
D.~Garreau and D.~Mardaoui.
\newblock What does lime really see in images?
\newblock In \emph{International Conference on Machine Learning}, 2021.

\bibitem[Goyal et~al.(2019)Goyal, Wu, Ernst, Batra, Parikh, and Lee]{pmlr-v97-goyal19a}
Y.~Goyal, Z.~Wu, J.~Ernst, D.~Batra, D.~Parikh, and S.~Lee.
\newblock Counterfactual visual explanations.
\newblock In K.~Chaudhuri and R.~Salakhutdinov, editors, \emph{Proceedings of the 36th International Conference on Machine Learning}, volume~97 of \emph{Proceedings of Machine Learning Research}, pages 2376--2384. PMLR, 09--15 Jun 2019.

\bibitem[He et~al.(2016)He, Zhang, Ren, and Sun]{DBLP:journals/corr/HeZRS15}
K.~He, X.~Zhang, S.~Ren, and J.~Sun.
\newblock Deep residual learning for image recognition.
\newblock In \emph{Proceedings of the IEEE Conference on Computer Vision and Pattern Recognition (CVPR)}, June 2016.

\bibitem[Hoyer et~al.(2019)Hoyer, Munoz, Katiyar, Khoreva, and Fischer]{10.5555/3454287.3454867}
L.~Hoyer, M.~Munoz, P.~Katiyar, A.~Khoreva, and V.~Fischer.
\newblock Grid saliency for context explanations of semantic segmentation.
\newblock In H.~Wallach, H.~Larochelle, A.~Beygelzimer, F.~d\textquotesingle Alch\'{e}-Buc, E.~Fox, and R.~Garnett, editors, \emph{Advances in Neural Information Processing Systems}, volume~32. Curran Associates, Inc., 2019.
\newblock URL \url{https://proceedings.neurips.cc/paper_files/paper/2019/file/6950aa02ae8613af620668146dd11840-Paper.pdf}.

\bibitem[Kapishnikov et~al.(2019)Kapishnikov, Bolukbasi, Vi{\'e}gas, and Terry]{kapishnikov2019xrai}
A.~Kapishnikov, T.~Bolukbasi, F.~Vi{\'e}gas, and M.~Terry.
\newblock Xrai: Better attributions through regions.
\newblock In \emph{Proceedings of the IEEE/CVF international conference on computer vision}, pages 4948--4957, 2019.

\bibitem[Ke et~al.(2022)Ke, Hwang, Guo, Wang, and Yu]{ke2022unsupervisedhierarchicalsemanticsegmentation}
T.-W. Ke, J.-J. Hwang, Y.~Guo, X.~Wang, and S.~X. Yu.
\newblock Unsupervised hierarchical semantic segmentation with multiview cosegmentation and clustering transformers.
\newblock In \emph{2022 IEEE/CVF Conference on Computer Vision and Pattern Recognition (CVPR)}, pages 2561--2571, 2022.
\newblock \doi{10.1109/CVPR52688.2022.00260}.

\bibitem[Khani et~al.(2024)Khani, Asgari, Sanghi, Amiri, and Hamarneh]{khani2024slime}
A.~Khani, S.~Asgari, A.~Sanghi, A.~M. Amiri, and G.~Hamarneh.
\newblock {SL}ime: Segment like me.
\newblock In \emph{The Twelfth International Conference on Learning Representations}, 2024.
\newblock URL \url{https://openreview.net/forum?id=7FeIRqCedv}.

\bibitem[Kim et~al.(2022)Kim, Meister, Ramaswamy, Fong, and Russakovsky]{kim2022hive}
S.~S. Kim, N.~Meister, V.~V. Ramaswamy, R.~Fong, and O.~Russakovsky.
\newblock Hive: Evaluating the human interpretability of visual explanations.
\newblock In \emph{European Conference on Computer Vision}, pages 280--298. Springer, 2022.

\bibitem[Kirillov et~al.(2023)Kirillov, Mintun, Ravi, Mao, Rolland, Gustafson, Xiao, Whitehead, Berg, Lo, Dollar, and Girshick]{kirillov2023segment}
A.~Kirillov, E.~Mintun, N.~Ravi, H.~Mao, C.~Rolland, L.~Gustafson, T.~Xiao, S.~Whitehead, A.~C. Berg, W.-Y. Lo, P.~Dollar, and R.~Girshick.
\newblock Segment anything.
\newblock In \emph{Proceedings of the IEEE/CVF International Conference on Computer Vision (ICCV)}, pages 4015--4026, October 2023.

\bibitem[Knab et~al.(2025)Knab, Marton, and Bartelt]{full_paper}
P.~Knab, S.~Marton, and C.~Bartelt.
\newblock Beyond pixels: Enhancing lime with hierarchical features and segmentation foundation models, 2025.
\newblock URL \url{https://arxiv.org/abs/2403.07733}.

\bibitem[Li et~al.(2022)Li, Zhou, Wang, Li, and Yang]{li2022deep}
L.~Li, T.~Zhou, W.~Wang, J.~Li, and Y.~Yang.
\newblock Deep hierarchical semantic segmentation.
\newblock In \emph{Proceedings of the IEEE/CVF Conference on Computer Vision and Pattern Recognition}, pages 1246--1257, 2022.

\bibitem[Lin et~al.(2014)Lin, Maire, Belongie, Hays, Perona, Ramanan, Doll{\'a}r, and Zitnick]{coco}
T.-Y. Lin, M.~Maire, S.~Belongie, J.~Hays, P.~Perona, D.~Ramanan, P.~Doll{\'a}r, and C.~L. Zitnick.
\newblock Microsoft coco: Common objects in context.
\newblock In D.~Fleet, T.~Pajdla, B.~Schiele, and T.~Tuytelaars, editors, \emph{Computer Vision -- ECCV 2014}, pages 740--755, Cham, 2014. Springer International Publishing.
\newblock ISBN 978-3-319-10602-1.

\bibitem[Linardatos et~al.(2021)Linardatos, Papastefanopoulos, and Kotsiantis]{e23010018}
P.~Linardatos, V.~Papastefanopoulos, and S.~Kotsiantis.
\newblock Explainable ai: A review of machine learning interpretability methods.
\newblock \emph{Entropy}, 23\penalty0 (1), 2021.
\newblock ISSN 1099-4300.
\newblock \doi{10.3390/e23010018}.
\newblock URL \url{https://www.mdpi.com/1099-4300/23/1/18}.

\bibitem[Liu et~al.(2022)Liu, Mao, Wu, Feichtenhofer, Darrell, and Xie]{DBLP:journals/corr/abs-2201-03545}
Z.~Liu, H.~Mao, C.-Y. Wu, C.~Feichtenhofer, T.~Darrell, and S.~Xie.
\newblock A convnet for the 2020s.
\newblock In \emph{Proceedings of the IEEE/CVF Conference on Computer Vision and Pattern Recognition (CVPR)}, pages 11976--11986, June 2022.

\bibitem[Lundberg and Lee(2017)]{shap}
S.~M. Lundberg and S.-I. Lee.
\newblock A unified approach to interpreting model predictions.
\newblock In \emph{Proceedings of the 31st International Conference on Neural Information Processing Systems}, NIPS'17, page 4768–4777, Red Hook, NY, USA, 2017. Curran Associates Inc.
\newblock ISBN 9781510860964.

\bibitem[Luo et~al.(2020)Luo, Cheng, Xu, Yu, Zong, Chen, and Zhang]{10.5555/3495724.3497370}
D.~Luo, W.~Cheng, D.~Xu, W.~Yu, B.~Zong, H.~Chen, and X.~Zhang.
\newblock Parameterized explainer for graph neural network.
\newblock In \emph{Proceedings of the 34th International Conference on Neural Information Processing Systems}, NIPS'20, Red Hook, NY, USA, 2020. Curran Associates Inc.
\newblock ISBN 9781713829546.

\bibitem[maintainers and contributors(2016)]{torchvision2016}
T.~maintainers and contributors.
\newblock Torchvision: Pytorch's computer vision library.
\newblock \url{https://github.com/pytorch/vision}, 2016.

\bibitem[Miller(2019)]{MILLER20191}
T.~Miller.
\newblock Explanation in artificial intelligence: Insights from the social sciences.
\newblock \emph{Artificial Intelligence}, 267:\penalty0 1--38, 2019.
\newblock ISSN 0004-3702.
\newblock \doi{https://doi.org/10.1016/j.artint.2018.07.007}.

\bibitem[Molnar et~al.(2022)Molnar, K{\"o}nig, Herbinger, Freiesleben, Dandl, Scholbeck, Casalicchio, Grosse-Wentrup, and Bischl]{pitfalls}
C.~Molnar, G.~K{\"o}nig, J.~Herbinger, T.~Freiesleben, S.~Dandl, C.~A. Scholbeck, G.~Casalicchio, M.~Grosse-Wentrup, and B.~Bischl.
\newblock General pitfalls of model-agnostic interpretation methods for machine learning models.
\newblock In A.~Holzinger, R.~Goebel, R.~Fong, T.~Moon, K.-R. M{\"u}ller, and W.~Samek, editors, \emph{xxAI - Beyond Explainable AI: International Workshop, Held in Conjunction with ICML 2020, July 18, 2020, Vienna, Austria, Revised and Extended Papers}, pages 39--68, Cham, 2022. Springer International Publishing.
\newblock ISBN 978-3-031-04083-2.
\newblock \doi{10.1007/978-3-031-04083-2_4}.

\bibitem[Montavon et~al.(2019)Montavon, Binder, Lapuschkin, Samek, and M{\"u}ller]{montavon2019layer}
G.~Montavon, A.~Binder, S.~Lapuschkin, W.~Samek, and K.-R. M{\"u}ller.
\newblock \emph{Layer-Wise Relevance Propagation: An Overview}, pages 193--209.
\newblock Springer International Publishing, Cham, 2019.
\newblock ISBN 978-3-030-28954-6.
\newblock \doi{10.1007/978-3-030-28954-6_10}.
\newblock URL \url{https://doi.org/10.1007/978-3-030-28954-6_10}.

\bibitem[Nauta et~al.(2023)Nauta, Trienes, Pathak, Nguyen, Peters, Schmitt, Schl\"{o}tterer, van Keulen, and Seifert]{anecdotal}
M.~Nauta, J.~Trienes, S.~Pathak, E.~Nguyen, M.~Peters, Y.~Schmitt, J.~Schl\"{o}tterer, M.~van Keulen, and C.~Seifert.
\newblock From anecdotal evidence to quantitative evaluation methods: A systematic review on evaluating explainable ai.
\newblock \emph{ACM Comput. Surv.}, 55\penalty0 (13s), jul 2023.
\newblock ISSN 0360-0300.
\newblock \doi{10.1145/3583558}.

\bibitem[Neubert and Protzel(2014)]{6976891}
P.~Neubert and P.~Protzel.
\newblock Compact watershed and preemptive slic: On improving trade-offs of superpixel segmentation algorithms.
\newblock In \emph{2014 22nd International Conference on Pattern Recognition}, pages 996--1001, 2014.
\newblock \doi{10.1109/ICPR.2014.181}.

\bibitem[Ng et~al.(2022)Ng, Abuwala, and Lim]{10082810}
C.~H. Ng, H.~S. Abuwala, and C.~H. Lim.
\newblock Towards more stable lime for explainable ai.
\newblock In \emph{2022 International Symposium on Intelligent Signal Processing and Communication Systems (ISPACS)}, pages 1--4, 2022.
\newblock \doi{10.1109/ISPACS57703.2022.10082810}.

\bibitem[Prasse et~al.(2023)Prasse, Jung, Bravo, Walter, and Keuper]{prasse2023towards}
K.~Prasse, S.~Jung, I.~B. Bravo, S.~Walter, and M.~Keuper.
\newblock Towards understanding climate change perceptions: A social media dataset.
\newblock In \emph{NeurIPS 2023 Workshop on Tackling Climate Change with Machine Learning}, 2023.
\newblock URL \url{https://www.climatechange.ai/papers/neurips2023/3}.

\bibitem[Radford et~al.(2021)Radford, Kim, Hallacy, Ramesh, Goh, Agarwal, Sastry, Askell, Mishkin, Clark, Krueger, and Sutskever]{radford2021learningtransferablevisualmodels}
A.~Radford, J.~W. Kim, C.~Hallacy, A.~Ramesh, G.~Goh, S.~Agarwal, G.~Sastry, A.~Askell, P.~Mishkin, J.~Clark, G.~Krueger, and I.~Sutskever.
\newblock Learning transferable visual models from natural language supervision, 2021.
\newblock URL \url{https://arxiv.org/abs/2103.00020}.

\bibitem[Ramamurthy et~al.(2020)Ramamurthy, Vinzamuri, Zhang, and Dhurandhar]{10.5555/3495724.3496225}
K.~N. Ramamurthy, B.~Vinzamuri, Y.~Zhang, and A.~Dhurandhar.
\newblock Model agnostic multilevel explanations.
\newblock In \emph{Proceedings of the 34th International Conference on Neural Information Processing Systems}, NIPS'20, Red Hook, NY, USA, 2020. Curran Associates Inc.
\newblock ISBN 9781713829546.

\bibitem[Ramesh et~al.(2021)Ramesh, Pavlov, Goh, Gray, Voss, Radford, Chen, and Sutskever]{DBLP:journals/corr/abs-2102-12092}
A.~Ramesh, M.~Pavlov, G.~Goh, S.~Gray, C.~Voss, A.~Radford, M.~Chen, and I.~Sutskever.
\newblock Zero-shot text-to-image generation.
\newblock In M.~Meila and T.~Zhang, editors, \emph{Proceedings of the 38th International Conference on Machine Learning}, volume 139 of \emph{Proceedings of Machine Learning Research}, pages 8821--8831. PMLR, 18--24 Jul 2021.
\newblock URL \url{https://proceedings.mlr.press/v139/ramesh21a.html}.

\bibitem[Rashid et~al.(2024)Rashid, Amparore, Ferrari, and Verda]{Rashid_Amparore_Ferrari_Verda_2024}
M.~Rashid, E.~G. Amparore, E.~Ferrari, and D.~Verda.
\newblock Using stratified sampling to improve lime image explanations.
\newblock \emph{Proceedings of the AAAI Conference on Artificial Intelligence}, 38\penalty0 (13):\penalty0 14785--14792, Mar. 2024.
\newblock \doi{10.1609/aaai.v38i13.29397}.

\bibitem[Ravi et~al.(2024)Ravi, Gabeur, Hu, Hu, Ryali, Ma, Khedr, Rädle, Rolland, Gustafson, Mintun, Pan, Alwala, Carion, Wu, Girshick, Dollár, and Feichtenhofer]{ravi2024sam2segmentimages}
N.~Ravi, V.~Gabeur, Y.-T. Hu, R.~Hu, C.~Ryali, T.~Ma, H.~Khedr, R.~Rädle, C.~Rolland, L.~Gustafson, E.~Mintun, J.~Pan, K.~V. Alwala, N.~Carion, C.-Y. Wu, R.~Girshick, P.~Dollár, and C.~Feichtenhofer.
\newblock Sam 2: Segment anything in images and videos, 2024.
\newblock URL \url{https://arxiv.org/abs/2408.00714}.

\bibitem[Ribeiro et~al.(2016)Ribeiro, Singh, and Guestrin]{lime}
M.~T. Ribeiro, S.~Singh, and C.~Guestrin.
\newblock "why should i trust you?": Explaining the predictions of any classifier.
\newblock In \emph{Proceedings of the 22nd ACM SIGKDD International Conference on Knowledge Discovery and Data Mining}, KDD '16, page 1135–1144, New York, NY, USA, 2016. Association for Computing Machinery.
\newblock ISBN 9781450342322.
\newblock \doi{10.1145/2939672.2939778}.
\newblock URL \url{https://doi.org/10.1145/2939672.2939778}.

\bibitem[Schallner et~al.(2020)Schallner, Rabold, Scholz, and Schmid]{superpixel_influence}
L.~Schallner, J.~Rabold, O.~Scholz, and U.~Schmid.
\newblock Effect of superpixel aggregation on explanations in lime -- a case study with biological data.
\newblock In P.~Cellier and K.~Driessens, editors, \emph{Machine Learning and Knowledge Discovery in Databases}, pages 147--158, Cham, 2020. Springer International Publishing.
\newblock ISBN 978-3-030-43823-4.

\bibitem[Schwab and Karlen(2019)]{10.5555/3454287.3455204}
P.~Schwab and W.~Karlen.
\newblock Cxplain: Causal explanations for model interpretation under uncertainty.
\newblock In H.~Wallach, H.~Larochelle, A.~Beygelzimer, F.~d\textquotesingle Alch\'{e}-Buc, E.~Fox, and R.~Garnett, editors, \emph{Advances in Neural Information Processing Systems}, volume~32. Curran Associates, Inc., 2019.

\bibitem[Sun et~al.(2023)Sun, Ma, Yuan, and Wang]{sun2023explain}
A.~Sun, P.~Ma, Y.~Yuan, and S.~Wang.
\newblock Explain any concept: Segment anything meets concept-based explanation.
\newblock In \emph{Thirty-seventh Conference on Neural Information Processing Systems}, 2023.
\newblock URL \url{https://openreview.net/forum?id=X6TBBsz9qi}.

\bibitem[Tan and Le(2019)]{tan2019efficientnet}
M.~Tan and Q.~Le.
\newblock Efficientnet: Rethinking model scaling for convolutional neural networks.
\newblock In \emph{International conference on machine learning}, pages 6105--6114. PMLR, 2019.

\bibitem[Tan et~al.(2024)Tan, Tian, and Li]{glime}
Z.~Tan, Y.~Tian, and J.~Li.
\newblock Glime: General, stable and local lime explanation.
\newblock \emph{Advances in Neural Information Processing Systems}, 36, 2024.

\bibitem[Wang et~al.(2017)Wang, Liu, Gao, Ma, and Soomro]{wang2017superpixel}
M.~Wang, X.~Liu, Y.~Gao, X.~Ma, and N.~Q. Soomro.
\newblock Superpixel segmentation: A benchmark.
\newblock \emph{Signal Processing: Image Communication}, 56:\penalty0 28--39, 2017.
\newblock ISSN 0923-5965.

\bibitem[Wang et~al.(2023)Wang, Li, Kallidromitis, Kato, Kozuka, and Darrell]{wang2024hierarchical}
X.~Wang, S.~Li, K.~Kallidromitis, Y.~Kato, K.~Kozuka, and T.~Darrell.
\newblock Hierarchical open-vocabulary universal image segmentation.
\newblock In A.~Oh, T.~Neumann, A.~Globerson, K.~Saenko, M.~Hardt, and S.~Levine, editors, \emph{Advances in Neural Information Processing Systems}, volume~36, pages 21429--21453. Curran Associates, Inc., 2023.

\bibitem[Zafar and Khan(2021)]{dlime}
M.~R. Zafar and N.~Khan.
\newblock Deterministic local interpretable model-agnostic explanations for stable explainability.
\newblock \emph{Machine Learning and Knowledge Extraction}, 3\penalty0 (3):\penalty0 525--541, 2021.
\newblock ISSN 2504-4990.
\newblock \doi{10.3390/make3030027}.
\newblock URL \url{https://www.mdpi.com/2504-4990/3/3/27}.

\bibitem[Zhang et~al.(2021)Zhang, Xu, Zou, Petrosian, and Krinkin]{9472817}
Y.~Zhang, F.~Xu, J.~Zou, O.~L. Petrosian, and K.~V. Krinkin.
\newblock Xai evaluation: Evaluating black-box model explanations for prediction.
\newblock In \emph{2021 II International Conference on Neural Networks and Neurotechnologies (NeuroNT)}, pages 13--16, 2021.
\newblock \doi{10.1109/NeuroNT53022.2021.9472817}.

\bibitem[Zhang et~al.(2023)Zhang, Gu, Song, Pan, Bai, and Zhao]{zhang2023xai}
Y.~Zhang, S.~Gu, J.~Song, B.~Pan, G.~Bai, and L.~Zhao.
\newblock Xai benchmark for visual explanation, 2023.

\bibitem[Zhao et~al.(2020)Zhao, Huang, Robu, and Flynn]{bayeslime}
X.~Zhao, X.~Huang, V.~Robu, and D.~Flynn.
\newblock Baylime: Bayesian local interpretable model-agnostic explanations.
\newblock In \emph{Conference on Uncertainty in Artificial Intelligence}, 2020.
\newblock URL \url{https://api.semanticscholar.org/CorpusID:227334656}.

\bibitem[Zhou et~al.(2021)Zhou, Hooker, and Wang]{slime}
Z.~Zhou, G.~Hooker, and F.~Wang.
\newblock S-lime: Stabilized-lime for model explanation.
\newblock In \emph{Proceedings of the 27th ACM SIGKDD Conference on Knowledge Discovery \& Data Mining}, KDD '21, page 2429–2438, New York, NY, USA, 2021. Association for Computing Machinery.
\newblock ISBN 9781450383325.

\end{thebibliography}

\newpage
\onecolumn
\appendix
\tableofcontents

\newpage
\section{Algorithms}
\subsection{DSEG-LIME algorithm}
\label{pseudocode}
We present the pseudocode of our DSEG-LIME framework in Algorithm \ref{alg:Hierarchical_LIME}. To construct the hierarchical segmentation within our framework, we start by calculating the overlaps between all segments in $\mathcal{S}$. We build a hierarchical graph using this overlap information through the following process.

\begin{algorithm}
\caption{DSEG-LIME framework pseudocode}\label{alg:Hierarchical_LIME}
\begin{algorithmic}[1]
\STATE {\bfseries Input:} $f$ (black-box model), $\zeta$ (segmentation function), $x$ (input instance), $g$ (interpretable model), $d$ (maximum depth), $hp$ (segmentation hyperparameters), $\theta$ (minimum segment size), $k$ (top segments to select)

\STATE \textbf{1. Initial segmentation:}
\STATE $\mathcal{S} \gets \zeta(x, hp)$ 

\STATE \textbf{2. Small cluster removal:}
\STATE $\mathcal{S'} \gets \{ s_i \in \mathcal{S} \,|\, \text{size}(s_i) \geq \theta \}$ 

\STATE \textbf{3. Hierarchical ordering:}
\STATE $\mathcal{H} \gets \text{BuildHierarchy}(\mathcal{S'})$ 

\FOR{$l \gets 1$ to $d$}
    \STATE \textbf{4. Empty space removal:}
    \IF{$l = 1$}
        \STATE $\mathcal{S}_l \gets \mathcal{H}[l]$ 
    \ELSE
        \STATE $\mathcal{S}_l \gets \{ s_i \in \mathcal{H}[l] \,|\, \text{parent}(s_i) \in top\_ids \}$ 
    \ENDIF

    \STATE $\mathcal{S}_l \gets \text{NearestNeighbor}(\mathcal{S}_l)$ 
    \STATE $Z \gets \text{Perturb}(x, \mathcal{S}_l)$ 
    \STATE $w \gets \text{Proximity}(Z, x)$ 
    \STATE $\text{preds} \gets f(z) \text{ for all } z \in Z$ 
    \STATE $g \gets \text{InitializeModel}(g)$ 
    \STATE $g \gets \text{Fit}(g, Z, \text{preds}, w)$ 
    \STATE $top\_ids \gets \{ \text{id}(s_i) \,|\, s_i \in \mathcal{S}_l, s_i \text{ is among top } k \text{ features in } g \}$ 
\ENDFOR

\STATE \textbf{Return} $g$ 
\end{algorithmic}
\end{algorithm}

First, we identify the top-level segments, which do not occur as subparts of any other segments. These segments serve as the highest-level nodes in the hierarchical graph. Starting from these top-level segments, we apply a top-down approach to identify child segments recursively. For each parent segment, we check for segments within it; these segments are designated as child nodes of the parent in the graph. Only the selected nodes are exclusively used to train the surrogate model.

This recursive process continues for each subsequent level, ensuring that every parent node encompasses its child nodes. The hierarchical graph thus formed represents the structural relationships between segments, where parent-child relationships indicate that child segments are complete parts of their respective parent segments. By constructing the hierarchy in this manner, we capture the nested structure of segments, which supports multi-level interpretability within the DSEG-LIME framework.

\newpage
\subsection{Hierarchical composition}
\label{hcompo}

The algorithm, designed for hierarchical composition, constructs a graph to represent the segments of a given instance, as shown in \Cref{alg:Hierarchical_LIME_SpecifiedOrder}. A key feature of this algorithm is the additional hyperparameter $t$, which determines when one segment is considered a subpart of another based on their overlap. Initially, the graph includes loops and redundant edges. To ensure it accurately represents the hierarchical relationships, the algorithm applies a series of predefined rules to tune the graph, removing unnecessary connections and simplifying its structure for the final representation.

\begin{algorithm}
\caption{DSEG-LIME Hierarchical Compostion}
\label{alg:Hierarchical_LIME_SpecifiedOrder}
\begin{algorithmic}[1]
\STATE {\bfseries Input:} $\mathcal{S'}$ (Segments), $t$ (Threshold)

\STATE \textbf{1. Build overlap matrix:}
\STATE Initialize overlap matrix $M$ of size $\|\mathcal{S'}\| \times \|\mathcal{S'}\|$ with zeros
\STATE Initialize graph $G$ with nodes corresponding to $\mathcal{S'}$
\FOR{$i \gets 1$ to $\|\mathcal{S'}\|$}
    \FOR{$j \gets i$ to $\|\mathcal{S'}\|$}  
        \STATE $M[i, j] \gets \frac{|\mathcal{S'}_i \cap \mathcal{S'}_j|}{|\mathcal{S'}_j|}$  
        \IF{$M[i, j] < t$} 
            \STATE $M[i, j] \gets 0$ 
        \ELSE
            \STATE Add edge $(i, j)$ to $G$ with weight $M[i, j]$ 
        \ENDIF
    \ENDFOR
\ENDFOR

\STATE \textbf{2. Remove loops in graph:}
\FOR{each node $v \in G$}
    \IF{edge $(v, v)$ exists}
        \STATE Remove edge $(v, v)$ 
    \ENDIF
\ENDFOR

\STATE \textbf{3. Prune graph:}
\FOR{each node $v \in G$}
    \STATE Find all paths from $v$ to other nodes 
    \FOR{each node $u$ reachable from $v$ via multiple paths}
        \STATE Identify the shortest path with the highest total weight 
        \STATE If multiple paths have the same weight, select the longest one 
        \STATE Remove all other paths from $v$ to $u$ 
    \ENDFOR
\ENDFOR

\STATE \textbf{4. Create dictionary from graph:}
\STATE Initialize dictionary $D$
\FOR{each node $v \in G$}
    \STATE $D[v] \gets \text{List of nodes connected to } v$ 
\ENDFOR

\STATE \textbf{Return} $D$ 
\end{algorithmic}
\end{algorithm}

\newpage
\section{Supplementary model evaluations} \label{model_evals}

\subsection{ResNet} \label{res}

In \Cref{tab:res_1}, we detail the quantitative results for the ResNet-101 model, comparing our evaluation with the criteria used for EfficientNetB4 under consistent hyperparameter settings. The review encompasses a comparative analysis with EfficientNetB3, emphasizing performance under contrastive conditions. The findings substantiate the results obtained from EfficientNet, demonstrating that the LIME techniques exhibit unpredictable behavior in the presence of model noise or prediction shuffling, despite varied segmentation strategies. This suggests an inherent randomness in the model explanations. Furthermore, the results indicate that all XAI methodologies displayed performance levels that were inferior to those of EfficientNetB4. However, it is significant to note that DSEG surpassed all other methods in terms of performance. 

\begin{table}[h]
  \caption{\textbf{Quantitative summary - classes ResNet-101.} The table presents the metrics consistently with those discussed for EfficientNet.
  }
\centering
\Large
\vspace{0.2cm}
\label{tab:res_1}
\resizebox{\columnwidth}{!}{%
\begin{tabular}{ll|rrrr|rrrr|rrrr|rrrr|rrrr}
\toprule
\multirow{2}{*}{Domain} &
  \multirow{2}{*}{Metric} &
  \multicolumn{4}{c}{DSEG} &
  \multicolumn{4}{c}{SLIC} &
  \multicolumn{4}{c}{QS} &
  \multicolumn{4}{c}{FS} &
  \multicolumn{4}{c}{WS} \\ \cmidrule(r){3-6} \cmidrule(r){7-10} \cmidrule(r){11-14} \cmidrule(r){15-18} \cmidrule(r){19-22} 
                                     &                       & L & S & G & B & L & S & G & B & L & S & G & B & L & S & G & B & L & S & G & B \\ 
\midrule
\multirow{3}{*}{Correctness}         & Random Model $\uparrow$         &85   &87  &\textbf{96}    & 92   & 85  & 85   & 86  &  80 &92 & 91  & 92  & 91 &  91  & 90   & 92&  90    & 81&  88&  92&91\\

                                     & Random Expl. $\uparrow$         &  95   &93    &97  & 94  & 94  & 92  & 94 &  91 & 94  & 94  & 92  &   95&  96 & 94  & 97  &97   & 91&  95 &  98 & \textbf{99}   \\
                                     
                                     & Single Deletion $\uparrow$      & 20   &19   &19   & \textbf{24}  & 10  &11  & 11  &  9 & 4  & 3  & 4  &   5 &  10 & 14  & 9  &10   & 9  & 9 &  14 &11 \\
\midrule
\multirow{2}{*}{\begin{tabular}[c]{@{}l@{}}Output \\ Completeness\end{tabular}} & Preservation $\uparrow$         &\textbf{48}   &39    &38  & 39  & 42  & 45  & 40  &  37 & 26  & 27  & 28  &   34 &  37 & 35  & 36  &36   & 29 &  38 &  39 &38  \\

                                     & Deletion $\uparrow$             &\textbf{56}   &50   & 50 & 50  & 40& 42  & 42  &  40 & 31  & 30  & 35  &   36&  39 & 32  & 35  &34   & 31  &  38 &  30 &41  \\
\midrule
\multirow{1}{*}{Consistency}         & Noise Preservation $\uparrow$
         & \textbf{45} & 39 & 42 & 36 & 38 & 40 & 41 & 35 & 26 & 24 & 23 & 26 & 28 & 30 & 30 & 30 & 28 & 37 & 32 & 33 \\
         & Noise Deletion $\uparrow$
         & 57 & \textbf{58} & 52 & 50 & 42 & 45 & 43 & 39 & 33 & 29 & 32 & 32 & 30 & 29 & 31 & 41 & 33 & 35 & 29 & 37 \\
\midrule
\multirow{2}{*}{Contrastivity}       & Preservation $\uparrow$& \textbf{45}   & 39   &39   & 43 & 36  & 35  & 36  &  35 & 31  & 37  & 30  & 40 &  42 & 43  & 43  &39   & 42  &41 &  44 &41 \\

                                     & Deletion $\uparrow$       & 47   &47   &47   & \textbf{48}  & 41  & 45  & 44  &  46 & 31  & 34  & 30  &  37&  33 & 34  & 34  &37   & 30  &  36 &  31 & 33  \\
\bottomrule
\end{tabular}
}%
\end{table}

\Cref{tab:res_2} presents further findings of ResNet. SLIME with DSEG yields the lowest AUC for incremental deletion, whereas Quickshift and Felzenszwalb show the highest. WS produces the smallest superpixels for compactness, contrasting with DSEG's larger ones. The stability analysis shows that all segmentations are almost at the same level, with QS being the best and GLIME the best-performing overall. Echoing EfficientNet's review, segmentation defines runtime, with DSEG being the most time-consuming. 

\begin{table}[h]
\caption{\textbf{Quantitative summary - numbers ResNet-101.} The table presents the metrics consistently with those discussed for EfficientNet.
  }
  \vspace{0.2cm}
\label{tab:res_2}
\Large
\resizebox{\columnwidth}{!}{%
\begin{tabular}{l|rrrr|rrrr|rrrr|rrrr|rrrr}
\toprule
\multirow{2}{*}{Metric} & \multicolumn{4}{c}{DSEG}   & \multicolumn{4}{c}{SLIC}  & \multicolumn{4}{c}{QS}    & \multicolumn{4}{c}{FS}    & \multicolumn{4}{c}{WS}    \\
\cmidrule(r){2-5} \cmidrule(r){6-9} \cmidrule(r){10-13} \cmidrule(r){14-17} \cmidrule(r){18-21}
                        & L    & S    & G    & B    & L    & S    & G    & B    & L    & S    & G    & B    & L    & S    & G    & B    & L    & S    & G    & B    \\
                        \midrule
Gini $\uparrow$ & 0.43 & 0.44 & \textbf{0.45} & 0.44 & 0.41 & 0.42 & 0.42 & 0.41 & 0.38 & 0.39 & 0.38 & 0.40 & 0.41 & 0.40 &0.41 & 0.39 & 0.41 & 0.41 & 0.42 & 0.40 \\
Incr. Deletion $\downarrow$         & 0.59 & \textbf{0.32} & 0.34 & 0.36 & 0.59 & 0.57 & 0.56 & 0.61 & 0.97 & 0.98 & 0.92 & 0.93 & 0.96 & 0.95 & 0.92 & 0.93 & 0.52 & 0.56 & 0.54 & 0.54 \\

Compactness $\downarrow$            & 0.22& 0.23 & 0.22 & 0.22 & 0.17 & 0.17 & 0.17 & 0.17 & 0.14 & 0.13 & 0.13 & 0.13 & 0.15 & 0.15 & 0.15 & 0.15 & 0.13 & 0.13 & \textbf{0.12} & 0.13 \\

Rep. Stability $\downarrow$         & .021 & .021 & .018 & .021 & .019 & .019& .016 & .019 & .018 & .018 & \textbf{.015} & .018 & .018 & .018 & .016 & .017 & .019 &  .019& .016 & .019 \\

Time $\downarrow$                   & 8.0 & 8.2 & 8.1 & 8.4 & 2.5 & 2.5 & \textbf{2.4} & 2.4 & 11.4 & 11.5 & 11.5 & 11.4 & 2.8 & 2.7 & 2.8 & 2.9 & 2.9 & 2.9 & 3.2 & 22\\
\bottomrule

\end{tabular}%
}
\end{table}

\newpage
\subsection{VisionTransformer} \label{vis_}

\Cref{tab:vit_1} provides the quantitative results for the VisionTransformer (ViT-384) model, employing settings identical to those used for EfficientNet and ResNet, with ViT processing input sizes of (384x384). The class-specific results within this table align closely with the performances recorded for the other models, further underscoring the effectiveness of DSEG. This consistency in DSEG performance is also evident in the data presented in \Cref{tab:vit_2}. 

We performed all experiments for ResNet and ViT with the same hyperparameters defined for EfficientNetB4. We would like to explicitly point out that the quantitative results could be improved by defining more appropriate hyperparameters for both DSEG and conventional segmentation methods, as no hyperparameter search was performed for a fair comparison.

\begin{table}[h]
  \caption{\textbf{Quantitative summary - classes ViT-384.} The table presents the metrics consistently with those discussed for EfficientNet.
  }
\centering
\Large
\vspace{0.2cm}
\label{tab:vit_1}
\resizebox{\columnwidth}{!}{%
\begin{tabular}{ll|rrrr|rrrr|rrrr|rrrr|rrrr}
\toprule
\multirow{2}{*}{Domain} &
  \multirow{2}{*}{Metric} &
  \multicolumn{4}{c}{DSEG} &
  \multicolumn{4}{c}{SLIC} &
  \multicolumn{4}{c}{QS} &
  \multicolumn{4}{c}{FS} &
  \multicolumn{4}{c}{WS} \\ \cmidrule(r){3-6} \cmidrule(r){7-10} \cmidrule(r){11-14} \cmidrule(r){15-18} \cmidrule(r){19-22} 
                                     &                       & L & S & G & B & L & S & G & B & L & S & G & B & L & S & G & B & L & S & G & B \\ 
\midrule
\multirow{3}{*}{Correctness}         
& Random Model $\uparrow$         &84   &85  &86  & 85   & 87  & 88  & 87  &  87 & 89  &89  &\textbf{90}  &  89&  88  & 87   & 86   &  86    & 86&  86&  87&86\\
                                     & Random Expl. $\uparrow$         &  93   &94    &96   & 95  & 97 & 95  & 94  &  96 & 94  & 94  & \textbf{99}  &   96&  95 & 95  & 98  &98   & 94&  94 &  91 & 95   \\
                                     
                                     & Single Deletion $\uparrow$      & 43   &43   &\textbf{44}    & 43  & 21  & 22  & 21  &  24 & 20  & 19  &18  &   17&  20& 20  & 22  &20   & 18  &  17 &  19 &20 \\
                                     
\midrule
\multirow{2}{*}{\begin{tabular}[c]{@{}l@{}}Output \\ Completeness\end{tabular}} & Preservation $\uparrow$         &\textbf{37}   &35  &35  &36  & 27  & 27   & 29  &  28 & 24  & 25  & 25&   25&  22 & 20  & 24  &20   & 23&  24 &  25 &25  \\

                                     & Deletion $\uparrow$             &82   &\textbf{84}   &\textbf{84}   & 83  & 74  & 74  & 73  &  73 & 62  & 61  & 62  &   61&  63 & 64  & 63  &66   & 67&  66 &  67 &66  \\
\midrule
\multirow{1}{*}{Consistency}         & Noise Preservation $\uparrow$
         & \textbf{32} & 30 & 29 & \textbf{32} & 27 & 28 & 29 & 27 & 23 & 21 & 22 & 23 & 22 & 21 & 21 & 22 & 23 & 25 & 24 & 26 \\
         & Noise Deletion $\uparrow$
         & 82 & 82 & 82 & \textbf{84} & 71 & 73 & 74 & 74 & 67 & 67 & 66 & 65 & 61 & 62 & 61 & 61 & 66 & 65 & 65 & 67 \\
\midrule
\multirow{2}{*}{Contrastivity}       & Preservation $\uparrow$& 63   &61    &63   & \textbf{64}  & 56  & 53  &  54 & 54  & 42  & 47  &   46&  43 &  59 & 58  & 60  & 59   & 46  &  50 &  49 & 47 \\

                                     & Deletion $\uparrow$       & 69   &68   &\textbf{72}   & 69  & 46  & 46  & 46  &  46 & 40  & 40  & 41  & 40 &  61 & 62  & 61  &61   & 42  &  42 &  43 & 42   \\
\bottomrule
\end{tabular}
}%
\end{table}

\begin{table}[h]
\caption{\textbf{Quantitative summary - numbers ViT-384.} The table presents the metrics consistently with those discussed for EfficientNet.
  }
\vspace{0.2cm}
\Large
\label{tab:vit_2}
\resizebox{\columnwidth}{!}{%
\begin{tabular}{l|rrrr|rrrr|rrrr|rrrr|rrrr}
\toprule
\multirow{2}{*}{Metric} & \multicolumn{4}{c}{DSEG}   & \multicolumn{4}{c}{SLIC}  & \multicolumn{4}{c}{QS}    & \multicolumn{4}{c}{FS}    & \multicolumn{4}{c}{WS}    \\
\cmidrule(r){2-5} \cmidrule(r){6-9} \cmidrule(r){10-13} \cmidrule(r){14-17} \cmidrule(r){18-21}
                        & L    & S    & G    & B    & L    & S    & G    & B    & L    & S    & G    & B    & L    & S    & G    & B    & L    & S    & G    & B    \\
                        \midrule
Gini $\uparrow$ & 0.52 & 0.52 & \textbf{0.53} & 0.52 & 0.48 & 0.49 & 0.48 & 0.47 & 0.40 & 0.41 & 0.42 & 0.40 & 0.44 & 0.42 &0.44 & 0.43 & 0.43 & 0.42 & 0.44 & 0.42 \\
Incr. Deletion $\downarrow$         & 1.00 & \textbf{0.46} &0.48 & 0.51 & 0.93 & 0.94 & 0.91 & 0.91 & 1.76 & 1.72& 1.71 & 1.72 & 1.64 & 1.62 & 1.54 & 1.59 & 1.02& 1.04 & 1.03 & 1.03 \\

Compactness $\downarrow$            & 0.20 & 0.20&0.20 & 0.19 & 0.15 & 0.14 & 0.15 & 0.15 & 0.12 & 0.12 & 0.12 & 0.12 & 0.14 & 0.14 & 0.14 & 0.14 &  \textbf{0.11} &  \textbf{0.11} &  \textbf{0.11} & 0.12 \\

Rep. Stability $\downarrow$         & \textbf{.013}  & \textbf{.013} & \textbf{.013} & \textbf{.013}  & .015 & .015& .016 & .015 & .018 & .018 & .019 & .018 & .017 & .017 & .017 & .016 & .017 &  .017& .018 & .017 \\

Time $\downarrow$                   & 6.6 & 6.5 & 6.6 & 6.6 & 3.3 & 3.2 & 3.3 & 3.3 & 12.2 & 12.1 & 12.3 & 12.3 & 2.6 & 2.6 &\textbf{ 2.5} & 2.7 & 3.6& 3.5 & 3.9& 3.6\\
\bottomrule

\end{tabular}%
}
\end{table}

\subsection{ConvNext} \label{ConvNext}

\Cref{tab:ConvNext} and \Cref{tab:ConvNext2} present the quantitative results pertaining to the ConvNext model \cite{DBLP:journals/corr/abs-2201-03545}, which is the last model that has been thoroughly evaluated in this study. The settings utilized in this evaluation remain consistent with those employed previously. 

\begin{table}[h]
  \caption{\textbf{Quantitative summary - classes ConvNext-224.} The table presents the metrics consistently with those discussed for EfficientNet.
  }
\centering
\Large
\vspace{0.2cm}
\label{tab:ConvNext}
\resizebox{\columnwidth}{!}{%
\begin{tabular}{ll|rrrr|rrrr|rrrr|rrrr|rrrr}
\toprule
\multirow{2}{*}{Domain} &
  \multirow{2}{*}{Metric} &
  \multicolumn{4}{c}{DSEG} &
  \multicolumn{4}{c}{SLIC} &
  \multicolumn{4}{c}{QS} &
  \multicolumn{4}{c}{FS} &
  \multicolumn{4}{c}{WS} \\ \cmidrule(r){3-6} \cmidrule(r){7-10} \cmidrule(r){11-14} \cmidrule(r){15-18} \cmidrule(r){19-22} 
                                     &                       & L & S & G & B & L & S & G & B & L & S & G & B & L & S & G & B & L & S & G & B \\ 
\midrule
\multirow{3}{*}{Correctness}         
& Random Model $\uparrow$         &92   &93  &\textbf{93}  & 92   & 81  & 80  & 81  &  81 & 79  &77  &79  &79&  82  & 82   & 81   &  82    & 83&  82&  82&82\\
                                     & Random Expl. $\uparrow$         &  89   &89    &90   & \textbf{94}  & 88 & 87  & 93  &  91 & \textbf{94}  & 91  & 90  &   93&  91 & 90  & 90  &88   & 89  &  86 &  92 & 91   \\
                                     
                                     & Single Deletion $\uparrow$      & 44   &\textbf{45}   &\textbf{45}    & 44  & 31  & 31  & 29  &  30 & 20  & 20  &21  &   19&  27& 26  & 26  &27   & 15  &  16 &  16 &16 \\
                                     
\midrule
\multirow{2}{*}{\begin{tabular}[c]{@{}l@{}}Output \\ Completeness\end{tabular}} & Preservation $\uparrow$         &\textbf{46}   &45  &\textbf{46}  &\textbf{46}  & 43  & 43   & 43  &  42 & 27  & 28  & 28&   28&  37 & 38  & 37  &36   & 35&  35 &  33 &35  \\

                                     & Deletion $\uparrow$             &\textbf{67}   &66   & 66  & 66  & 64  & 66  & 63  &  63 & 62  & 61  & 55  &   55&  55 & 55  & 55  &56   & 53&  54 &  54 &54  \\
\midrule
\multirow{1}{*}{Consistency}         & Noise Preservation $\uparrow$
         & 40 & \textbf{47} & \textbf{47} & \textbf{47} & 39 & 40 & 38 & 41 & 36 & 35 & 38 & 37 & 37 & 38 & 38 & 40 & 40 & 39 & 40 & 38 \\
         & Noise Deletion $\uparrow$
         & 68 & \textbf{70} & \textbf{70} & 69 & 63 & 63 & 64 & 62 & 56 & 56 & 56 & 55 & 57 &56 & 56 & 58 & 58 & 57 & 59 & 58 \\
\midrule
\multirow{2}{*}{Contrastivity}       & Preservation $\uparrow$& 58   &\textbf{62}    &61   & \textbf{62}  & 59  & 58  &  60 & 59  & 50  & 50  &   49&  46 &  51 & 53  & 52  & 51   & 55  &  56 &  53 & 54 \\

                                     & Deletion $\uparrow$       & \textbf{58}   &\textbf{58}   &\textbf{58}   & \textbf{58}  & 47  & 47  & 47  &  47& 44  & 43  & 44  & 43 &  37 & 37  & 37  &37   & 46  &  46 &  46 & 47   \\
\bottomrule
\end{tabular}
}%
\end{table}

\begin{table}[h]
\caption{\textbf{Quantitative summary - numbers ConvNext-224.} The table presents the metrics consistently with those discussed for EfficientNet.
  }
\vspace{0.2cm}
\Large
\label{tab:ConvNext2}
\resizebox{\columnwidth}{!}{%
\begin{tabular}{l|rrrr|rrrr|rrrr|rrrr|rrrr}
\toprule
\multirow{2}{*}{Metric} & \multicolumn{4}{c}{DSEG}   & \multicolumn{4}{c}{SLIC}  & \multicolumn{4}{c}{QS}    & \multicolumn{4}{c}{FS}    & \multicolumn{4}{c}{WS}    \\
\cmidrule(r){2-5} \cmidrule(r){6-9} \cmidrule(r){10-13} \cmidrule(r){14-17} \cmidrule(r){18-21}
                        & L    & S    & G    & B    & L    & S    & G    & B    & L    & S    & G    & B    & L    & S    & G    & B    & L    & S    & G    & B    \\
                        \midrule
Gini $\uparrow$ & 0.48 & 0.49 & 0.49 & 0.49 & \textbf{0.51} & 0.50 & \textbf{0.51} & \textbf{0.51} & 0.43 & 0.44 & 0.45 & 0.44 & 0.47 & 0.47 &0.48 & 0.49 & 0.48 & 0.48 & 0.48 & 0.48 \\
Incr. Deletion $\downarrow$         & 0.87 & \textbf{0.35} &0.36 & 0.37 & 0.61 & 0.60 & 0.60 & 0.60 & 1.13 & 1.14& 1.14 & 1.13 & 1.10 & 1.10 & 1.10 & 1.10 & 0.65& 0.64 & 0.65 & 0.65 \\

Compactness $\downarrow$            & 0.23 & 0.23&0.23 & 0.23 & 0.19 & 0.19 & 0.19 & 0.19 & 0.15 & 0.15 & \textbf{0.14 }& 0.15 & 0.16 & 0.16 & 0.16 & 0.16 &  0.16 &   0.16 &  0.16 & 0.16 \\

Rep. Stability $\downarrow$         & .012  & .012 & .013 & .012  & \textbf{.011} & \textbf{.011}& .012 & .012 & .012 & .012 & .013 & .012 & .012 & .012 & .013 & .012 & \textbf{.011} &  \textbf{.011}& \textbf{.011} & \textbf{.011}\\

Time $\downarrow$                   & 3.6 & 3.4 & 3.5 & 3.6 & 1.0 &\textbf{ 0.9} & 1.0 & 1.0 & 2.8 & 2.5 & 2.5 & 2.9 & 1.5 & 1.4 &1.5 & 1.5 & \textbf{0.9}& \textbf{0.9}& \textbf{0.9}& 1.0\\
\bottomrule

\end{tabular}%
}
\end{table}

\subsection{DSEG compared to EAC} \label{eac_sec}
We conducted additional experiments with Explain Any Concept (EAC) \citep{sun2023explain}, performing the same quantitative experiments as for DSEG, but with a reduced dataset comprising 50 images to optimize computational time (\Cref{app_data}). We began our evaluation by noting that EAC, unlike DSEG-LIME, cannot be applied to arbitrary models, which is a significant drawback of their method and prevents comprehensive comparisons. Thus, we compared our approach against LIME and EAC, in explaining ResNet. The results are listed in \Cref{eac_num} and \Cref{eac_num_2}.

\begin{table}[h]
  \caption{\textbf{Quantitative summary - classes EAC.}This table presents the metrics in line with the previous evaluations, focusing on ResNet performance for DSEG and other segmentation techniques in comparison to EAC.}
  \vspace{0.2cm}
\centering
\label{eac_num}
\resizebox{\columnwidth}{!}{
\Large
\begin{tabular}{ll|r|r|r|r|r|r|r}
\toprule
\multirow{1}{*}{Domain} &
  \multirow{1}{*}{Metric} &
   \multicolumn{1}{c}{LIME-DSEG} 
  & \multicolumn{1}{c}{LIME-SLIC}
   & \multicolumn{1}{c}{LIME-QS} 
   & \multicolumn{1}{c}{LIME-FS}
   & \multicolumn{1}{c}{LIME-WS}
   & \multicolumn{1}{c}{EAC} \\ 
\midrule
\multirow{3}{*}{Correctness}         
& Random Model $\uparrow$         &45   &42&44& 40&\textbf{46}&45 \\
                                     & Random Expl. $\uparrow$         &  \textbf{49}   &45    &45 &46&44  & 48 \\
                                     
                                     & Single Deletion $\uparrow$      & \textbf{10}   &7   &4 &7&8  & 1 \\
                                     
\midrule
\multirow{2}{*}{\begin{tabular}[c]{@{}l@{}}Output \\ Completeness\end{tabular}} & Preservation $\uparrow$         &20   &21   &12 &15&19 & \textbf{35}\\

                                     & Deletion $\uparrow$             &27  &18   &20  &18&18  & \textbf{38}  \\
\midrule
\multirow{1}{*}{Consistency}         & Noise Stability $\uparrow$      & 20  & 17    & 13 &12& 18 & \textbf{34} \\

\midrule
\multirow{2}{*}{Contrastivity}       & Preservation $\uparrow$& 17    &13    &19 &19& 21 & \textbf{31} \\

                                     & Deletion $\uparrow$       & 25   &23   &24  &22&18 & \textbf{35}  \\
\bottomrule
\end{tabular}
}
\end{table}

We observe that EAC quantitatively outperforms DSEG in certain cases. However, the results indicate that DSEG shows marked improvement as the number of samples increases, ultimately achieving comparable computation times. Moreover, it is expected that EAC performs better with ResNet, as it is specifically designed to leverage the model’s internal representations. The main drawback of EAC, however, is its lack of general applicability, as it cannot be used across all model architectures.

\begin{table}[h]
\caption{\textbf{Quantitative summary - numbers EAC.} This table presents the metrics in line with the previous evaluations, focusing on ResNet performance for DSEG and other segmentation techniques in comparison to EAC.}
\label{eac_num_2}
\vskip 0.15in
\centering
\footnotesize
\begin{tabular}{l|r|r|r|r|r|r|r}
\toprule
\multirow{1}{*}{Metric} &
  \multicolumn{1}{c}{LIME-DSEG} 
  & \multicolumn{1}{c}{LIME-SLIC}
   & \multicolumn{1}{c}{LIME-QS} 
   & \multicolumn{1}{c}{LIME-FS}
   & \multicolumn{1}{c}{LIME-WS}
   & \multicolumn{1}{c}{EAC} \\ 
\midrule

Incr. Deletion $\downarrow$         & 0.54 & 0.55 & 0.90& 0.85&0.50&\textbf{0.01} \\

Compactness $\downarrow$            & 0.25 & 0.16 &0.14& 0.16&0.13&\textbf{ 0.11} \\

Rep. Stability $\downarrow$         & .021 & .019& .018 & .018&.017&\textbf{.002} \\

Time $\downarrow$                   & 8.0 & \textbf{2.8} & 12.6 &2.9&3.5& 326.7 \\
\bottomrule

\end{tabular}%

\end{table}
\newpage
\subsection{DSEG within SLICE} \label{slice_chapt}

\Cref{tab:experimental_results} reports the results for the DSEG framework applied to the SLICE \cite{bora2024slice} approach and conducts a comparative evaluation against Quickshift (QS) in LIME.
The experiments were carried out on the minimized dataset consisting of 50 images in total. Each image was processed for 200 steps and repeated 8 times. The primary evaluation metric used was the average rank similarity \textit{ (p = 0.3)}, which focused on positive and negative segments and ensured comparability by employing the minimum segment count of both approaches. Furthermore, a \textit{ sign flip analysis} was performed, measuring the number of segments that changed signs during the computation. This analysis considered all segments produced by each technique. 

\begin{table}[h]
\centering
\caption{\textbf{DSEG and QS in SLICE.} The Average Rank Similarity shows high similarity for both approaches, while the Sign Flip metric indicates that DSEG produces fewer segment sign changes.}
\label{tab:experimental_results}
\vskip 0.15in
\begin{tabular}{lcc}
\toprule
\textbf{Metric} & \textbf{DSEG (pos, neg)} & \textbf{QS (pos, neg)} \\
\midrule
Average Rank Similarity (p=0.3) & 0.969347 / 0.971652 & 0.970756 / 0.970384 \\
Sign Flip (mean) & 2.608696 & 31.288043 \\
\bottomrule
\end{tabular}
\end{table}

DSEG and QS demonstrate comparable performance in rank similarity. However, DSEG markedly surpasses QS in the Sign Flip metric. This superior performance is attributable to QS producing a greater number of segments, whereas DSEG prioritizes generating meaningful segments that align more closely with the foundational model. These findings underscore DSEG’s advantage in necessitating less hyperparameter tuning and yielding more robust segmentation compared to alternative methodologies.

\subsection{DETR within DSEG} \label{detr}

In \Cref{tab:detr_c} and \Cref{tab:detr_n}, 
we conducted the DETR experiments within LIME. Based on prior results, we assess its performance by contrasting it with SLIC within the LIME framework, also utilizing the same condensed dataset comprising 50 images. Both experiments were configured with identical parameters, and DETR was implemented for basic panoptic segmentation. 

\begin{table}[h]
  \caption{\textbf{Quantitative summary - classes DETR.} The table showcases metrics for EfficientNetB4, and DETR within DSEG. Results reported pertain solely to integrating DSEG and SLIC within the scope of the LIME frameworks examined.}
\vskip 0.15in
\centering
\label{tab:detr_c}
\footnotesize
\begin{tabular}{ll|rrrr|rrrr}
\toprule
\multirow{2}{*}{Domain} &
  \multirow{2}{*}{Metric} &
  \multicolumn{4}{c}{DSEG} 
   & \multicolumn{4}{c}{SLIC} \\ \cmidrule(r){3-6} \cmidrule(r){7-10}
                                     &                       & L & S & G & B & L & S & G & B\\ 
\midrule
\multirow{3}{*}{Correctness}         
& Random Model $\uparrow$         &\textbf{32}&\textbf{32}&\textbf{32}& \textbf{32} & 30  & 30  & 30  &  30  \\
                                     & Random Expl. $\uparrow$         &  29   &37    &38   & 40 &38  & \textbf{45}  & 39  &  38  \\
                                     
                                     & Single Deletion $\uparrow$      & \textbf{36}   &\textbf{36}   &35   & \textbf{36} & 18  & 17  & 21  &  21 \\
                                     
\midrule
\multirow{2}{*}{\begin{tabular}[c]{@{}l@{}}Output \\ Completeness\end{tabular}} & Preservation $\uparrow$         &\textbf{43}   &42   &42  &42 & 37   &  35   & 35   &  35  \\

                                     & Deletion $\uparrow$             &34  &34    &\textbf{35}   & 34 & 21 & 21  & 21  &  21 \\
\midrule
\multirow{1}{*}{Consistency}         & Noise Stability $\uparrow$      & \textbf{40}  & \textbf{40}    & 39    & 39  & 35  & 36  & 36  &  36  \\

\midrule
\multirow{2}{*}{Contrastivity}       & Preservation $\uparrow$& \textbf{39}    &37    &36   & 36 & 28  & 28  & 27  &  28 \\

                                     & Deletion $\uparrow$       & \textbf{35}   &34   &33   & 32 & 23  & 24  & 24   &  24 \\
\bottomrule
\end{tabular}
\end{table}

\begin{figure}[h]
\centering
\includegraphics[width=0.9\textwidth]{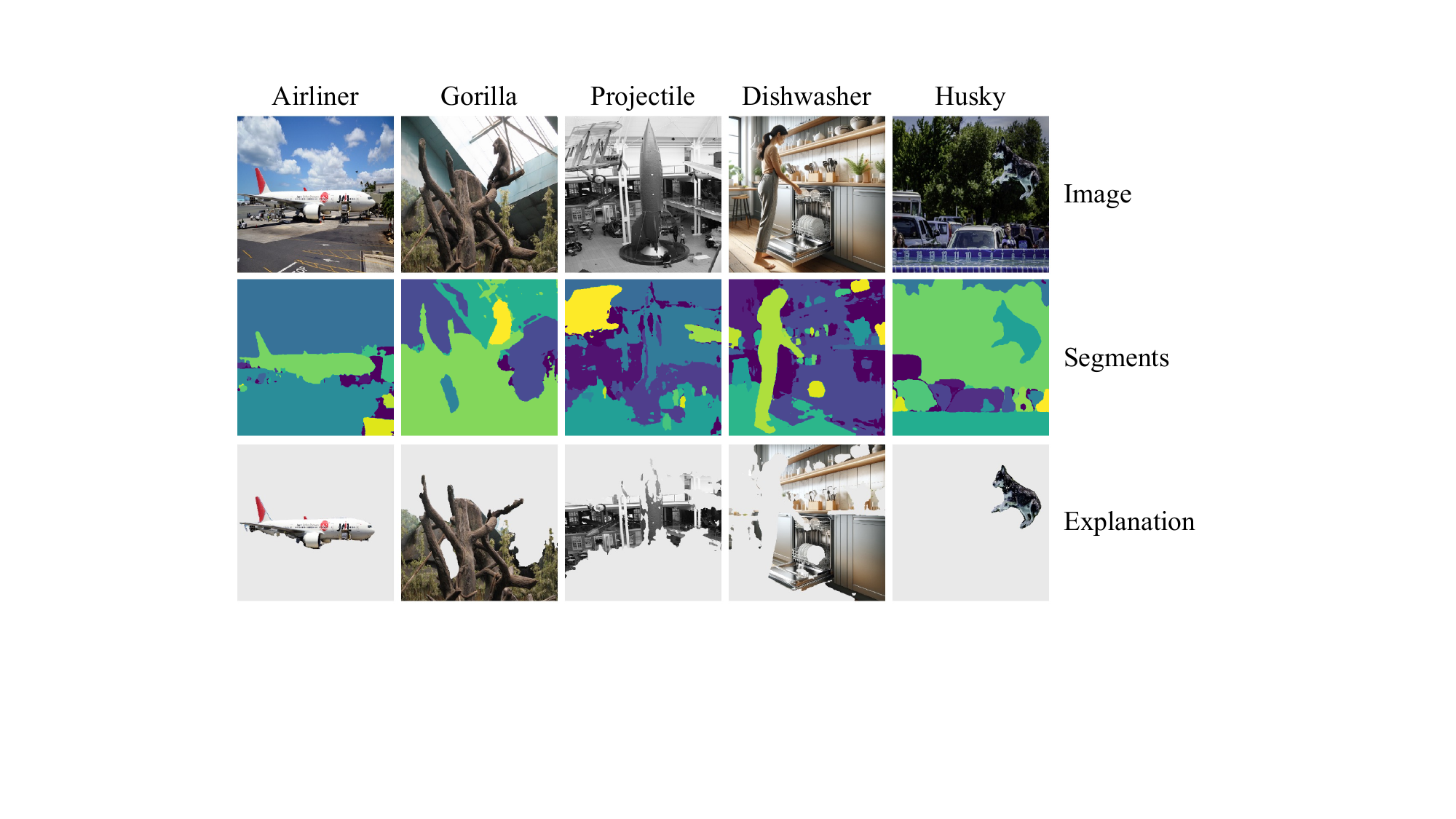}
\caption{\textbf{DETR within DSEG.} The visualization displays five instances with classes from the ImageNet dataset. Each image includes the prediction by EfficientNetB4 as its headline, the segmentation map of DETR, and the corresponding explanation by DETR within LIME.}
\label{fig:detr}
\end{figure}

\begin{table}[h]
\caption{\textbf{Quantitative summary - numbers DETR.} The table showcases the numeric values in the same manner as in \Cref{tab:detr_c} but for numeric values.}
\vskip 0.15in
\centering
\label{tab:detr_n}
\footnotesize
\begin{tabular}{l|rrrr|rrrr}
\toprule
\multirow{2}{*}{Metric} & \multicolumn{4}{c}{DSEG}  & \multicolumn{4}{c}{SLIC}    \\
\cmidrule(r){2-5}  \cmidrule(r){6-9}
                        & L    & S    & G    & B  & L    & S    & G    & B  \\
                        \midrule
Incr. Deletion $\downarrow$         &0.64 & 0.34 & 0.37& \textbf{0.25} & 0.68 & 0.70 & 0.75 & 0.69 \\

Compactness $\downarrow$            & 0.34 & 0.34 &0.34& 0.34& 0.15 & \textbf{0.14} & 0.15 & 0.15\\

Rep. Stability $\downarrow$         & .008 & .008& .008 & \textbf{.007}  & .010& .010& .011 & .010 \\

Time $\downarrow$                   & 23.6 & \textbf{22.0} & 24.4 & 23.5 & 22.9 & 24.5 & 27.6 & 25.6\\
\bottomrule
\end{tabular}
\end{table}
\newpage
DETR demonstrates superior performance on the dataset compared to the LIME variants utilizing SLIC. Despite its efficacy, the segmentation quality of DETR was generally inferior to that of SAM, as evidenced by less compact explanations. 
This observation is further supported by the examples in \Cref{fig:detr}. The visualizations reveal that DETR often segments images in ways that do not align with typical human-recognizable concepts, highlighting a potential limitation in its practical utility for generating explanatory segments. Moreover, DETR does not support the construction of a segmentation hierarchy, lacking the ability to produce finer and coarser segments, which diminishes its flexibility compared to methods such as SAM.

\newpage
\subsection{SAM 2 within DSEG} 
\label{sam_2}

In the main paper, we conducted experiments using SAM 1. In this part, we integrate SAM 2 \citep{ravi2024sam2segmentimages} with the 'hiera\_l' backbone into the DSEG framework, applying a 0.8 stability score threshold. We also report the results for EfficientNetB4 on the dataset comprising 50 images in \Cref{tab:sam20_c} and \Cref{tab:sam20_n}.

\begin{table}[ht]
  \caption{\textbf{Quantitative summary - classes SAM 2.} This table presents the metrics in line with those discussed for EfficientNet in the main paper, but displays only the results for SLIC for the sake of simplicity.}
  \vskip 0.15in
\centering
\label{tab:sam20_c}
\footnotesize
\begin{tabular}{ll|rrrr|rrrr}
\toprule
\multirow{2}{*}{Domain} &
  \multirow{2}{*}{Metric} &
  \multicolumn{4}{c}{DSEG} 
   & \multicolumn{4}{c}{SLIC} \\ 
   \cmidrule(r){3-6} \cmidrule(r){7-10}
                                     &                       & L & S & G & B & L & S & G & B\\ 
\midrule
\multirow{3}{*}{Correctness}         
& Random Model $\uparrow$         &\textbf{38}&\textbf{38}&\textbf{38}& \textbf{38} & 30  & 30  & 30  &  30  \\
                                     & Random Expl. $\uparrow$         &  40   &44    &43   & 44 &38  & \textbf{45}  & 39  &  38  \\
                                     
                                     & Single Deletion $\uparrow$      & 27   &27   &\textbf{28}   & \textbf{28} & 18  & 17  & 21  &  21 \\
                                     
\midrule
\multirow{2}{*}{\begin{tabular}[c]{@{}l@{}}Output \\ Completeness\end{tabular}} & Preservation $\uparrow$         &\textbf{39}   &34   &35  &34 & 37  &  35   & 35   &  35  \\

                                     & Deletion $\uparrow$             &\textbf{34}  &30    &30   & 30 & 21 & 21  & 21  &  21 \\
\midrule
\multirow{1}{*}{Consistency}         & Noise Stability $\uparrow$      & \textbf{38}  & \textbf{38}     & \textbf{38}     & 37  & 35  & 36  & 36  &  36  \\

\midrule
\multirow{2}{*}{Contrastivity}       & Preservation $\uparrow$& \textbf{31}    &28    &29   & 30 & 28  & 28  & 27  &  28 \\

                                     & Deletion $\uparrow$       & \textbf{35}   &30   &31   & 31 & 23  & 24   &24   & 24 \\
\bottomrule
\end{tabular}
\end{table}

As both tables demonstrate, DSEG-LIME consistently outperforms other methods and surpasses DSEG with SAM 1 across most metrics, delivering superior results. It effectively segments images into more meaningful regions, particularly in cases where SAM 1 faced challenges, reinforcing the conclusions of the SAM 2 technical report.

\begin{table}[ht]
\caption{\textbf{Quantitative summary - numbers SAM 2.} This table presents the metrics in line with those discussed for EfficientNet in the main paper, but displays only the results for SLIC for the sake of simplicity.}
\vskip 0.15in
\centering
\label{tab:sam20_n}
\footnotesize
\begin{tabular}{l|rrrr|rrrr}
\toprule
\multirow{2}{*}{Metric} & \multicolumn{4}{c}{DSEG}  & \multicolumn{4}{c}{SLIC}    \\
\cmidrule(r){2-5}  \cmidrule(r){6-9}
                        & L    & S    & G    & B  & L    & S    & G    & B  \\
                        \midrule

Incr. Deletion $\downarrow$         & 0.98 & \textbf{0.36} & \textbf{0.36}& 0.39 & 0.68 & 0.70 & 0.75 & 0.69 \\

Compactness $\downarrow$            & 0.16 & 0.16 &0.17& 0.17& 0.15 & \textbf{0.14} & 0.15 & 0.15\\

Rep. Stability $\downarrow$         & .011 & \textbf{.010}& .011 & \textbf{.010}  & \textbf{.010}& \textbf{.010}& .011 & \textbf{.010} \\

Time $\downarrow$                   & 19.1 & 18.9 & 18.7 & 19.3 & 14.7 & \textbf{14.6} & 14.8 &  14.8\\
\bottomrule

\end{tabular}
\end{table}

However, since the experiments were conducted on different hardware, the computation times vary. Here, we report the time for the SLIC variant of LIME, but similar to previous experiments, the times for other LIME variants with SLIC are expected to be comparable to those of standard LIME. As a result, DSEG with SAM 2 is slightly slower due to the additional segmentation process.

\begin{figure}[h]
\centering
\includegraphics[width=0.9\linewidth]{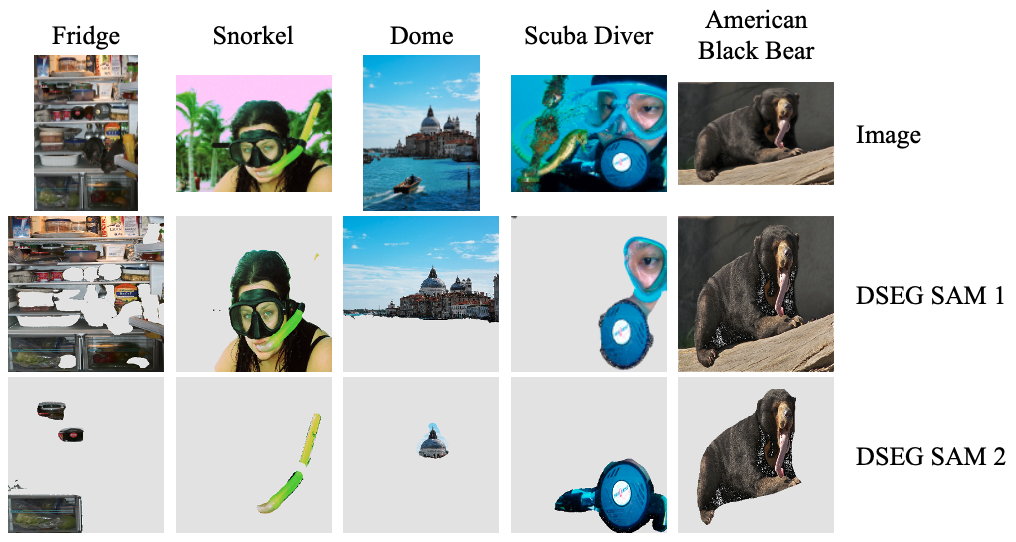}
\caption{\textbf{Comparison DSEG with SAM 1 and SAM 2.} Exemplary images with explanations generated by SAM 1 and SAM 2 within DSEG, illustrating how the updated SAM improves segment utilization for DSEG.}
\label{fig:SAM_2}
\end{figure}

\textbf{Exemplary explanations.}
\Cref{fig:SAM_2} presents explanations generated by DSEG using both SAM 1 and SAM 2, highlighting cases where the newer version of SAM enables DSEG to produce more meaningful and interpretable explanations. Each image includes explanations for the predicted class from EfficientNetB4. While SAM 2 shows improved segmentation in these examples, similar results can be obtained with SAM 1 by appropriately adjusting the hyperparameters for automatic mask generation.

\subsection{EfficientNetB4 with depth of two} \label{depth2}

In \Cref{tab:d2_c} and \Cref{tab:d2_n}, we present the quantitative comparison between DSEG-LIME ($d = 2$) using EfficientNetB4 and SLIC, as reported in the main paper. The hyperparameter settings were consistent across the evaluations, except for compactness. We established a minimum threshold of 0.05 for values to mitigate the impact of poor segmentation performance, which often resulted in too small segments. Additional segments were utilized to meet this criterion for scenarios with suboptimal segmentation. 
However, this compactness constraint was not applied to DSEG with depth two since its hierarchical approach naturally yields smaller and more detailed explanations, evident in \Cref{tab:d2_n}. 
The hierarchical segmentation of $d = 2$ slightly impacts stability, yet the method continues to generate meaningful explanations, as indicated by other metrics.
Although our method demonstrated robust performance, it required additional time because the feature attribution process was conducted twice.

\begin{table}[ht]
  \caption{\textbf{Quantitative summary - classes depth two.} The table showcases metrics for EfficientNetB4, specifically at a finer concept granularity; the hierarchical segmentation tree has $d = 2$. Results reported pertain solely to integrating DSEG and SLIC within the scope of the LIME frameworks examined.}
  \vskip 0.15in
\centering
\vskip 0.15in
\label{tab:d2_c}
\footnotesize
\begin{tabular}{ll|rrrr|rrrr}
\toprule
\multirow{2}{*}{Domain} &
  \multirow{2}{*}{Metric} &
  \multicolumn{4}{c}{DSEG} 
   & \multicolumn{4}{c}{SLIC} \\ 
   \cmidrule(r){3-6} \cmidrule(r){7-10}
                                     &                       & L & S & G & B & L & S & G & B\\ 
\midrule
\multirow{3}{*}{Correctness}         
& Random Model $\uparrow$         &62&68&\textbf{69}& \textbf{69}& 68  & 67  & 68  &  68  \\
                                     & Random Expl. $\uparrow$         &  87   &87    &87   & \textbf{91} &81  & 79  & 80  &  84  \\
                                     
                                     & Single Deletion $\uparrow$      & 38   &40   &\textbf{45}   & 43 & 36  & 34  & 33  &  34 \\
                                     
\midrule
\multirow{2}{*}{\begin{tabular}[c]{@{}l@{}}Output \\ Completeness\end{tabular}} & Preservation $\uparrow$         &63   &70   &68  &68 & 74   &  74   & \textbf{75}   &  74  \\

                                     & Deletion $\uparrow$             &45  &49    &50   & \textbf{52} & 39 & 39  & 39  &  40 \\
\midrule
\multirow{2}{*}{Consistency}         & Noise Preservation $\uparrow$      & 65  & 63& 63    & 64  & \textbf{72}  & \textbf{72}  & \textbf{72}  &  \textbf{72}  \\
                                    & Noise Deletion $\uparrow$      & \textbf{54}  & 52    & 53    & 52  & 40  & 41  & 40  &  40  \\

\midrule
\multirow{2}{*}{Contrastivity}       & Preservation $\uparrow$& 51  & 53& 52    & 53 & 54  & 54  & \textbf{55}  &  \textbf{55} \\

                                     & Deletion $\uparrow$       &\textbf{50}  & 46    & 49    & 48& 49  & \textbf{50}   & 48   &  \textbf{50} \\
\bottomrule
\end{tabular}
\end{table}

\begin{table}[ht]
\caption{\textbf{Quantitative summary - numbers depth two.} The table showcases the numeric values in the same manner as in \Cref{tab:d2_c} but for numeric values.}
\vskip 0.15in
\centering
\label{tab:d2_n}
\footnotesize
\begin{tabular}{l|rrrr|rrrr}
\toprule
\multirow{2}{*}{\textbf{Metric}} & \multicolumn{4}{c|}{\textbf{DSEG}}  & \multicolumn{4}{c}{\textbf{SLIC}} \\
        \cmidrule(lr){2-5} \cmidrule(lr){6-9}
                        & L    & S    & G    & B  & L    & S    & G    & B  \\
                        \midrule
Gini $\uparrow$ & 0.42 & 0.43 & 0.42 & 0.49  & 0.49  & 0.50  & 0.50&0.50\\

Incr. Deletion $\downarrow$         & 1.29 & 0.85 & \textbf{0.80}& 1.10 & 0.81 & 0.82 & 0.81 & 0.82 \\

Compactness $\downarrow$            & 0.16 & 0.17 &0.17& 0.16& \textbf{0.15} & \textbf{0.15} & \textbf{0.15} & \textbf{0.15}\\

Rep. Stability $\downarrow$         & .014 & .015& .016 & .015  & \textbf{.011}& \textbf{.011}& .012 & .012 \\

Time $\downarrow$                   & 47.6 & 52.5 & 53.4 & 52.8 & \textbf{22.9} & 24.5 & 27.6 & 25.6\\
\bottomrule

\end{tabular}
\end{table}

\begin{figure}[h]
\centering
\includegraphics[width=0.6\linewidth]{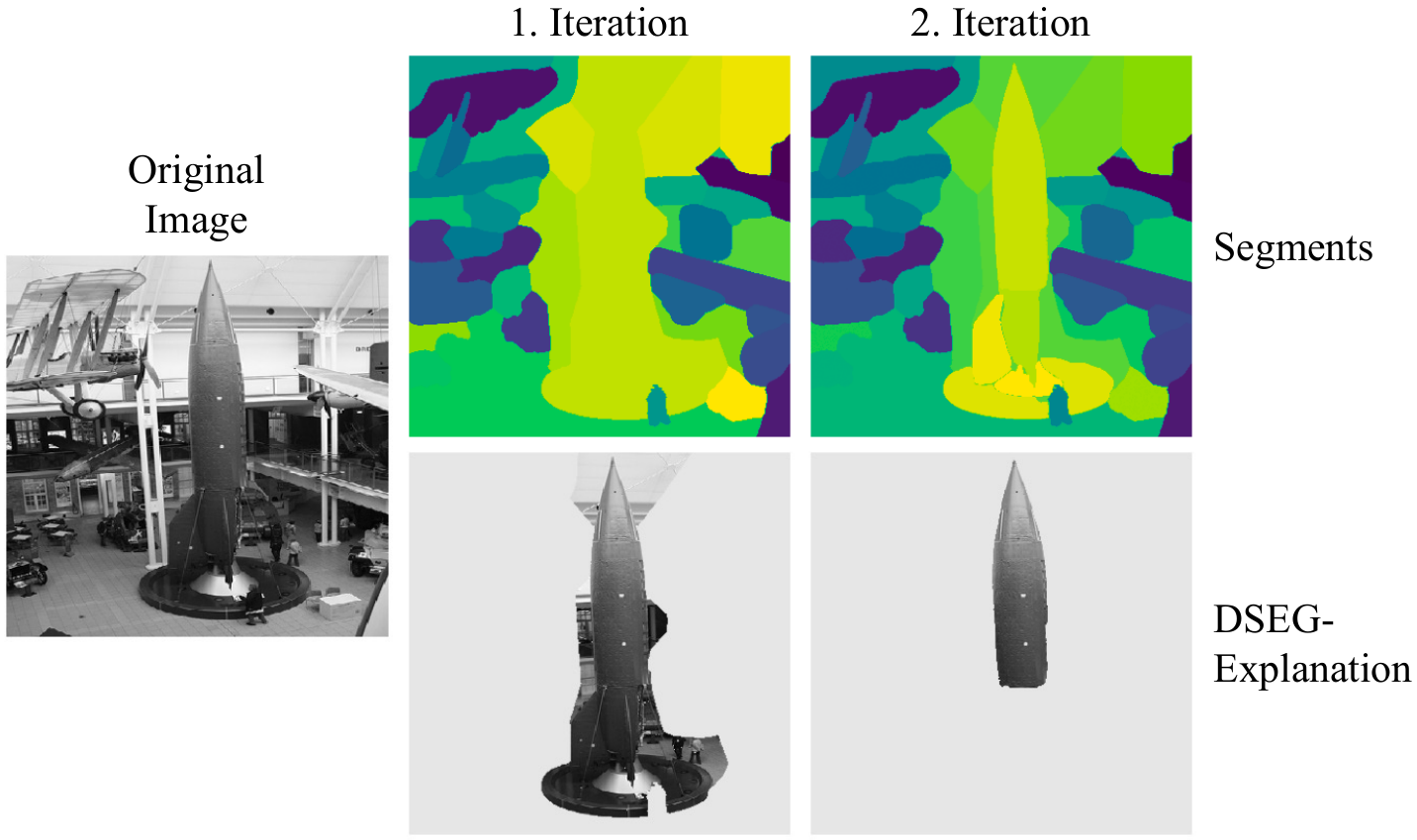}
\caption{\textbf{2nd iteration of DSEG-LIME.} Visualizing DSEG's explanations of a projectile. It includes the first iteration's explanation along with its corresponding segmentation map. Additionally, similar details are provided for the second iteration procedure, highlighting the upper part of a projectile as an explanation.}
\label{fig:projectile}
\end{figure}

\newpage
\textbf{Exemplary explanations.} 
In \Cref{fig:projectile}, an instance is explained with DSEG and $d = 2$, showing a black-and-white image of a projectile. Here, we see the corresponding explanation for each stage, starting with the first iteration, with the corresponding segmentation map. In the second iteration, we see the segment representing the projectile split into its finer segments - the children nodes of the parent node - with the corresponding explanation below.

\begin{figure}[h]
\centering
\includegraphics[width=0.75\linewidth]{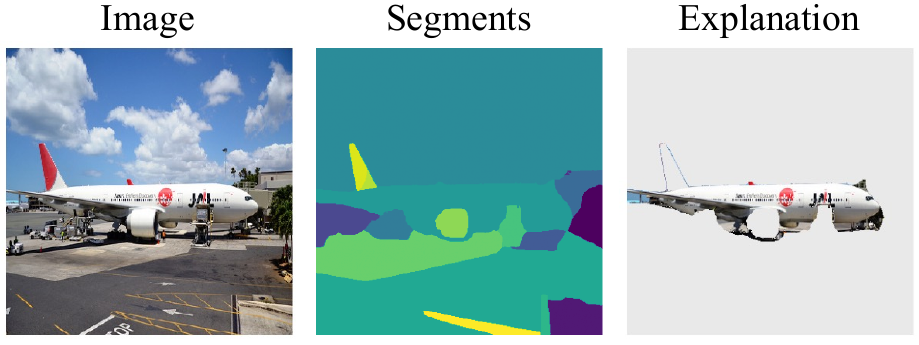}
\caption{\textbf{Airliner explanation with depth two.} The same example as in \Cref{fig:pipeline_images} but with segmentation hierarchy of two for the explanation. This example includes the children nodes of the most significant parent node in the segmentation map for feature importance calculation.}
\label{second_iter}
\end{figure}

\newpage
\textbf{Case study.} We examine the case presented in \Cref{fig:pipeline_images}, where DSEG initially segments the image into various layers with overlapping features, establishing a segmentation hierarchy through composition. In the first iteration, LIME focuses solely on the segments just beneath the root node - the parent segments that cannot be merged into broader concepts. From this segmentation map, LIME determines the feature importance scores, identifying the airplane as the most crucial element in the image.
In the subsequent iteration, illustrated in \Cref{second_iter}, DSEG generates an additional segmentation map that further divides the airplane into finer components for detailed analysis. The explanation in this phase emphasizes the airplane's body, suggesting that this concept of the 'Airliner' is most significant. 

\newpage
\section{Further experiments with DSEG} \label{further_experiments}

\subsection{Ablation study} \label{ablation_study}
\textbf{Automatic mask generation.} Ablating DSEG using only the automatic masks produced by foundation models presents notable challenges, particularly when applying methods like LIME. These automatically generated masks often result in many segments, some of which may overlap or leave parts of the image unsegmented. Such inconsistencies make a direct integration with LIME impractical. Consequently, our ablation in this setting is limited to comparing different foundation models for mask generation. We conducted experiments with SAM, SAM2, and DETR to evaluate their respective impact on DSEG’s performance.

\begin{figure}[h]
\centering
\includegraphics[width=0.8\linewidth]{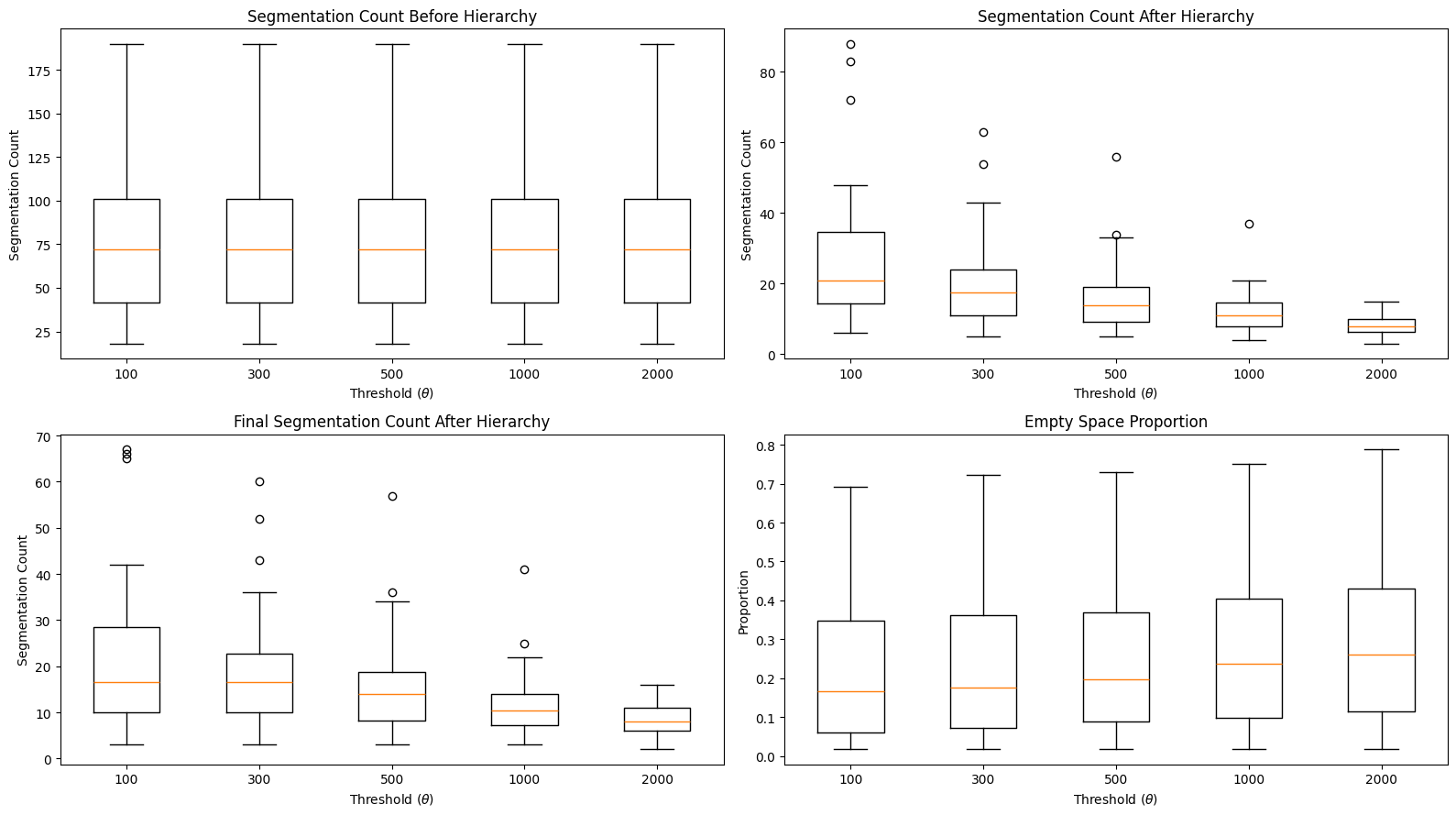}
\caption{\textbf{Ablation study.} Here we present the interquartile range (IQR) of segmentation counts at different stages of the DSEG process (before hierarchy, after hierarchy, and final segmentation) and the proportion of empty space across various threshold values for segment size removal (denoted by $\theta$).}
\label{fig:ablation}
\end{figure}

\textbf{Small cluster removal.} For the ablation study, we examine how the number of segments evolves across different stages of the segmentation process as we vary the threshold for removing segments smaller than the hyperparameter $\theta$ (with values [100, 300, 500, 1000, 2000]). Additionally, we assess the behavior of empty spaces within the segmented regions across all images in the dataset. The analysis focuses on three key points: the number of segments immediately after the initial automated segmentation, after hierarchical sorting, and after the removal of undersized segments, following the complete DSEG approach. The empty space is evaluated before it is filled with adjacent segments. A comprehensive overview of the metrics for these steps is presented in \Cref{fig:ablation}.

Higher thresholds lead to fewer segments being retained. This trend is visible in the segmentation counts before the hierarchy, after the hierarchy, and in the final segmentation. For instance, at $\theta = 100$, a higher number of segments is preserved, whereas at $\theta = 2000$, the segmentation count drops significantly due to the removal of smaller segments.
Additionally, the proportion of empty space consistently increases with larger $\theta$ values. This occurs because as more small segments are removed, more unassigned or empty regions appear before being filled by adjacent segments. The increase in empty space proportion is most pronounced at higher thresholds, such as $\theta = 1000$ and $\theta = 2000$.
In summary, the analysis highlights the expected trade-off between preserving smaller segments and controlling the amount of empty space. Lower thresholds result in more granular segmentation, while higher thresholds reduce the segmentation complexity at the expense of increased empty regions. Based on this trade-off, a threshold of $\theta = 500$ was selected for the experiments in this paper, as it strikes a balance between retaining meaningful segmentation detail and minimizing empty space.

\textbf{Hierarchical ordering.}
The effect of hierarchical ordering is already addressed through both qualitative and quantitative analyses across different depths of the explanation process. This hierarchical structure ensures no overlapping segments occur—a necessary condition for reliably computing feature attributions. As a result, an additional technical ablation isolating hierarchical ordering is not required.

\textbf{Empty space removal.}
The presence of unsegmented regions within an image is a well-known issue in the segmentation domain. One potential solution for adapting LIME to this scenario is to assign these unsegmented areas to a dedicated segment—e.g., a “segment zero.” However, this approach introduces the challenge that such unsegmented areas are often non-contiguous and scattered across the image. From a user perspective, this fragmented behavior is undesirable and could negatively affect interpretability. To address this, we opted for a simple yet practical solution: reassigning unsegmented pixels to the nearest segmented region using a nearest-neighbor approach. While effective for our purposes, this strategy opens avenues for future work, where more sophisticated treatments of unsegmented areas could be explored.

\subsection{Medical scenario with depth of three}
\label{med_3}

\begin{figure}[h] \centering 
\includegraphics[width=0.95\linewidth]{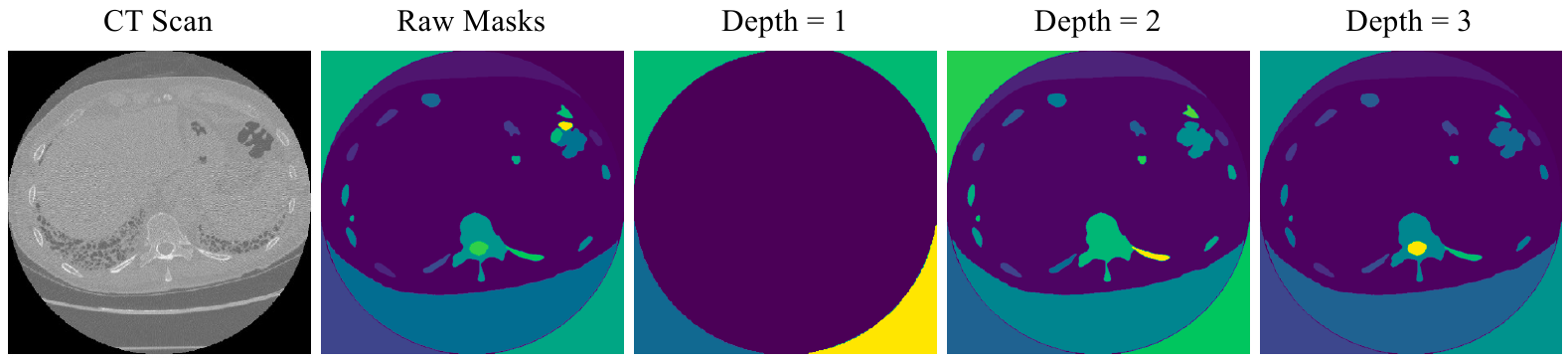} 
\caption{\textbf{Medical use case.} Demonstration of DSEG on a medical use case using CT scan imagery. From left to right: the original CT image, the raw segmentation mask, and hierarchical segmentations generated by DSEG at depths 1, 2, and 3. Each subsequent depth introduces finer segmentation details important for explanation. This showcases the general motivation for hierarchical explanations and the applicability of DSEG in medical imaging contexts.}
\label{d_3_health}
\end{figure}

We demonstrate the application of our framework on a medical use case, processing CT imagery of the human body, as shown in \Cref{d_3_health}. In this example, we present the original image alongside the raw segmentation mask and DSEG’s hierarchical compositions at depths 1 to 3, where each level introduces progressively finer segments relevant for explanation.

For this demonstration, we utilize the basic SAM variant. However, improved segmentation quality could be achieved by using fine-tuned models specifically adapted to the medical domain. It is important to note that we do not provide a detailed explanation of a specific instance here; rather, we aim to illustrate the general motivation behind DSEG. Thus, the presented hierarchy serves demonstration purposes but highlights the feasibility and applicability of our framework when paired with appropriate segmentation models for medical data.

Furthermore, we show that the segmentation hierarchy can extend beyond depth 2, as previously evaluated in \Cref{depth2}. As a final remark, if a user requests an explanation deeper than the maximum available hierarchy (determined by the most activated nodes), the explanation terminates due to the inherent limitation in segmentation depth.

\subsection{Explaining wrong classification} \label{wrong_class}

Here, we explore how DSEG can aid in explaining a model's misclassification. Unlike the previous analysis in \Cref{quanti}, where metrics were assessed under simulations involving a model with randomized weights (Random Model) or random predictions (Random Expl.), this case focuses on a real misclassification by EfficientNetB4, free from external manipulation. This allows for a more genuine examination of DSEG's ability to explain incorrect classifications under normal operating conditions.

\begin{figure}[h] \centering 
\includegraphics[width=0.6\linewidth]{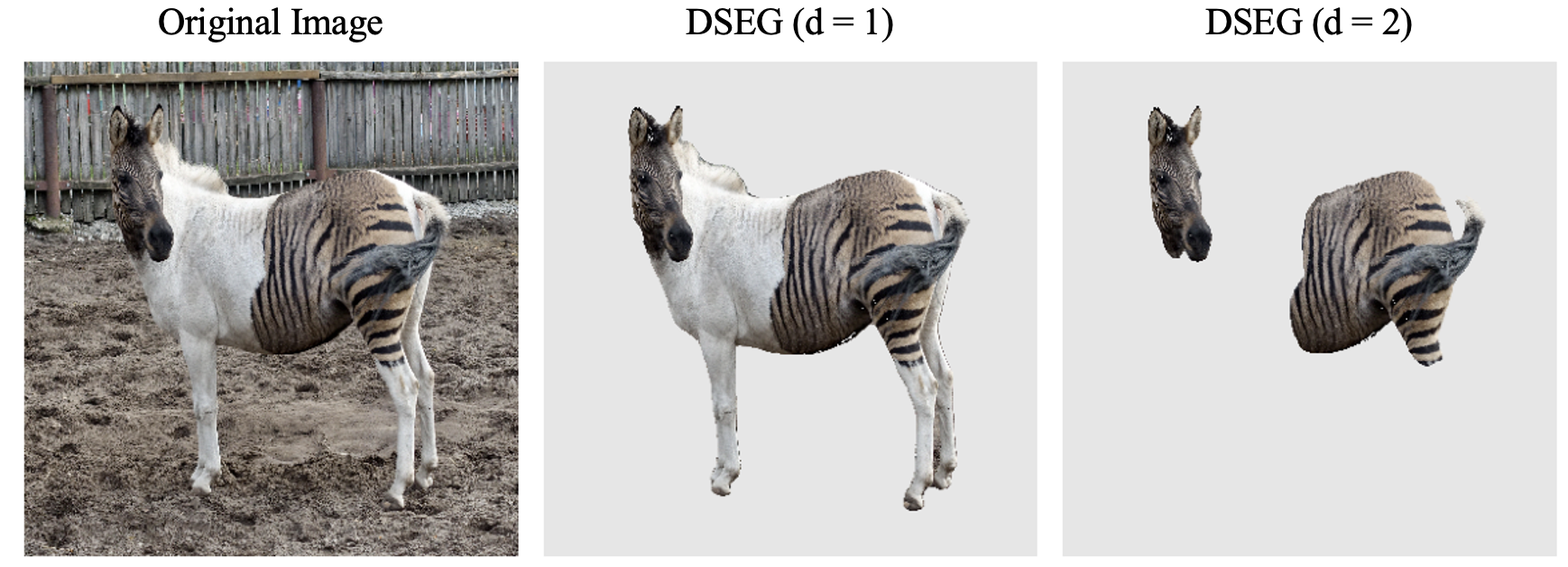} 
\caption{\textbf{Misclassification example.} The image depicts a hybrid of a horse and zebra that EfficientNetB4 classifies as a zebra with $p = 0.17$. DSEG-LIME, with a depth of one, highlights the entire animal, offering a broad explanation. Meanwhile, DSEG at depth two pinpoints specific zebra-like patterns that influence the model’s prediction. This suggests that the model is fixating on particular visual features associated with zebras, explaining its erroneous classification.} 
\label{fig_zebra}
\end{figure}

\Cref{fig_zebra} shows an image of a hybrid between a horse (sorrel) and a zebra, where EfficientNetB4 can recognize both animals but does not contain the hybrid class. We explore why EfficientNetB4 assigns the highest probability to the zebra class rather than the sorrel. Although this is not strictly a misclassification, it simulates a similar situation and provides insight into why the model favors the zebra label over the sorrel. This analysis helps us understand the model's decision-making process in cases where it prioritizes specific features associated with one class over another.

\subsection{Stability of explanations} \label{app_stab}

\begin{figure}[h]
\centering
\includegraphics[width=0.8\linewidth]{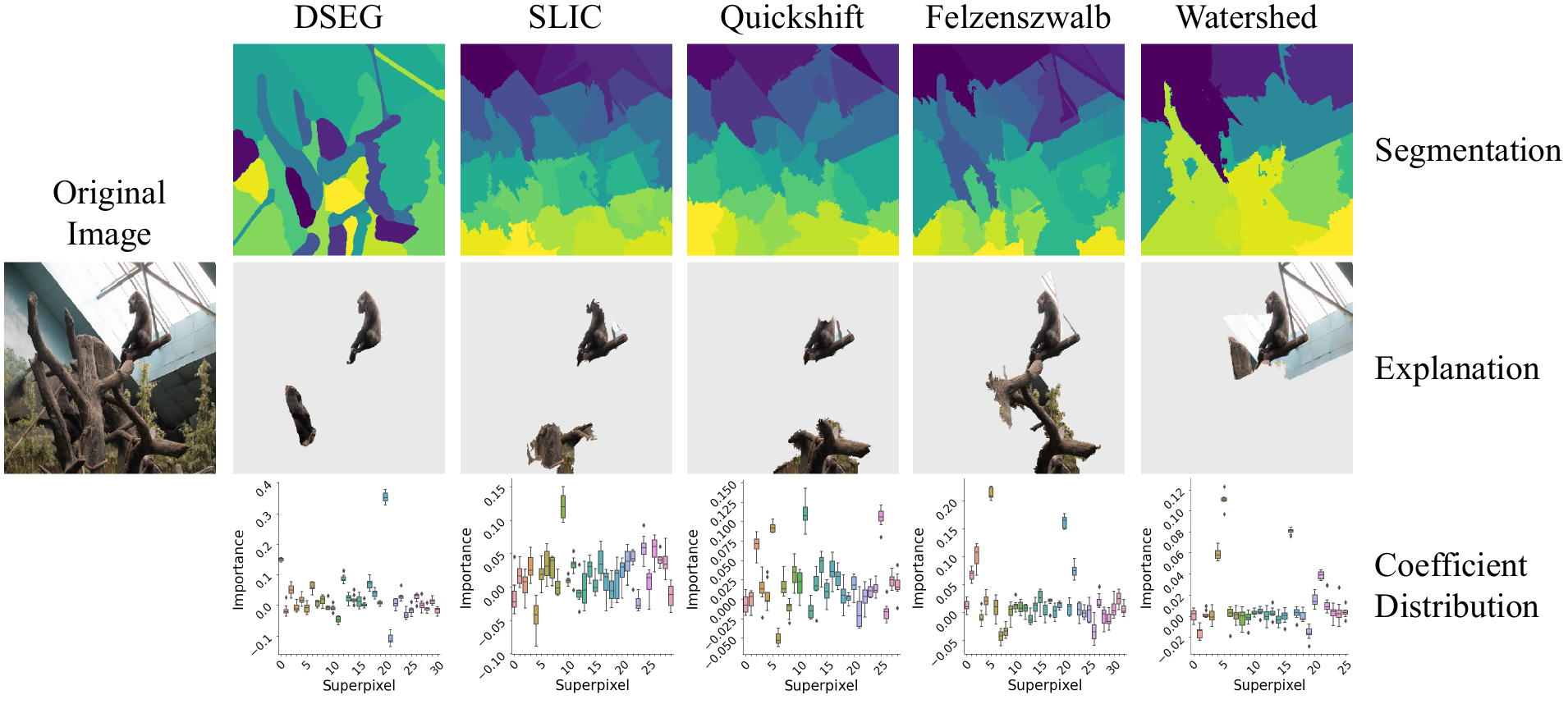}
\caption{\textbf{Segmentation stability.} Illustrating a comparison between DSEG and other segmentation techniques applied in LIME, all utilizing an identical number of samples. DSEG exhibits greater stability compared to other segmentation techniques. Notably, the DSEG explanation distinctly highlights the segment representing a gorilla as the most definitive.}
\label{fig:stability}
\end{figure}

The stability of imagery explanations using LIME can be linked to the quality of feature segments, as illustrated in \Cref{fig:stability}. This figure presents the segmentation maps generated by various techniques alongside their explanations and coefficient distributions, displayed through an IQR plot over eight runs. Notably, the DSEG technique divides the image into meaningful segments; the gorilla segment, as predicted by the EfficientNetB4 model, is distinctly visible and sharply defined. In contrast, other techniques also identify the gorilla, but less clearly, showing significant variance in their coefficient distributions. Watershed, while more stable than others, achieves this through overly broad segmentation, creating many large and a few small segments. These findings align with our quantitative evaluation and the described experimental setup.

\subsection{Zero-shot classification explanation} \label{clip_}

In this section, we demonstrate the versatility of DSEG-LIME by applying it to a different dataset and classification task. Specifically, we replicate the zero-shot classification approach described in \citep{prasse2023towards} using CLIP \citep{radford2021learningtransferablevisualmodels} for the animal super-category. Since DSEG-LIME maintains model-agnostic properties, it remains applicable to zero- and few-shot classification models without modification.

\begin{figure}[h] \centering \includegraphics[width=0.7\textwidth]{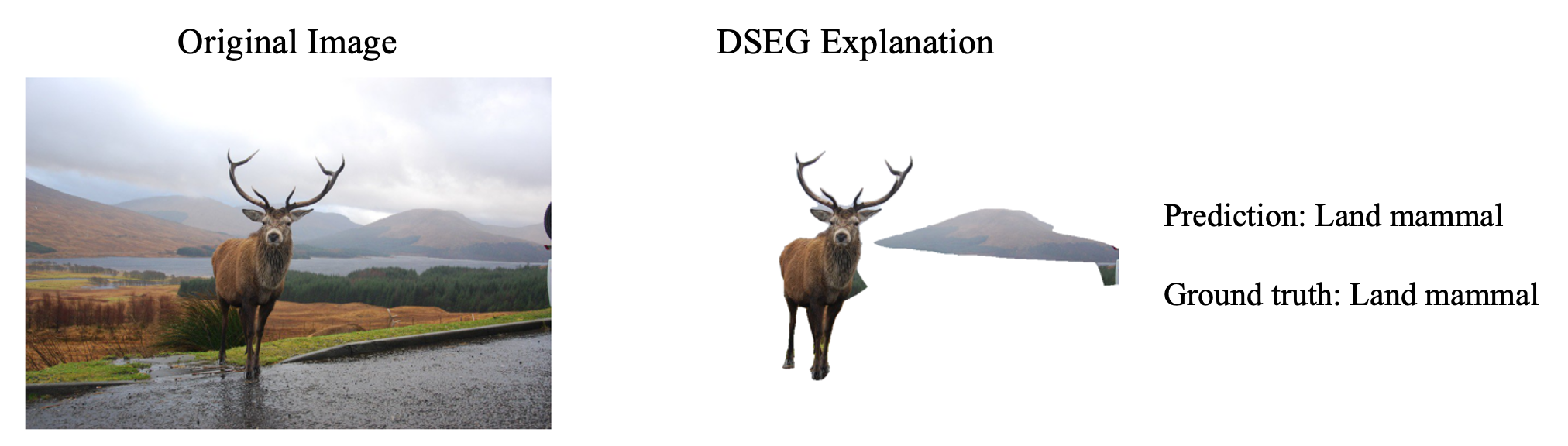} \caption{\textbf{DSEG-LIME explanation for CLIP.} This figure illustrates an image processed by CLIP for zero-shot classification into animal categories. The model correctly predicted the class as "Land mammal." DSEG-LIME highlights the two most important features influencing the classification, with the presence of the deer being the most significant.} \label{clip_fig} \end{figure}

\Cref{clip_fig} presents an illustrative example from the dataset, where the task is to classify an image into the animal category. The predicted and ground-truth class for the image is 'Land mammal'. As shown by DSEG-LIME's explanation, the model's decision is primarily influenced by the presence of a deer in the foreground and a mountain in the background, which contribute to the overall classification.

\newpage
\subsection{Exemplary limitation of DSEG} \label{fail_dseg}
\begin{figure}[h]
\centering
\includegraphics[width=0.6\textwidth]{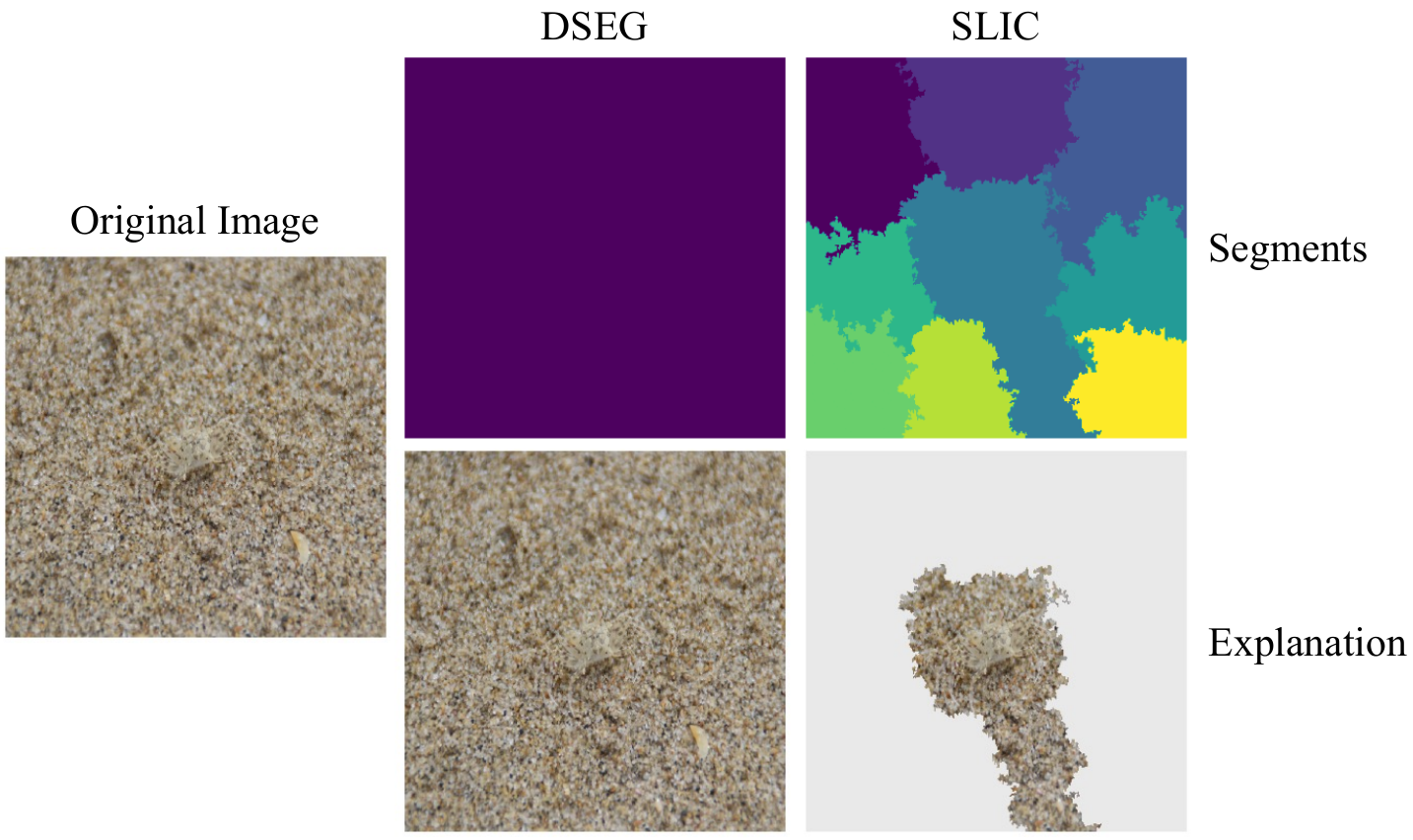}
\caption{\textbf{DSEG fails.} Demonstrating
a scenario where DSEG fails to generate
meaningful features for explanations (the
the whole image is one segment, in contrast
to SLIC. The image shows a crab, which
the model classifies as a ’hermit crab’ (p
= 0.17), highlighting the effectiveness of
SLIC in this context compared to the limitations of DSEG.}
\label{fig:Fails}
\end{figure}

The example in \Cref{fig:Fails} shows a complex case of a hermit crab in front of sand, which is hardly detectable. Here, SAM fails to segment the image into meaningful segments, a known issue in the community \citep{khani2024slime}. In contrast, SLIC can generate segments; thus, LIME can produce an explanation that does not show a complete image. However, to mitigate this issue, one could use a segmentation model that has better performance in such scenes, such as \cite{chen2023sam}.

\newpage
\subsection{Feature attribution maps} \label{feature_maps}
\begin{figure}[h!]
\centering
\includegraphics[width=0.7\linewidth]{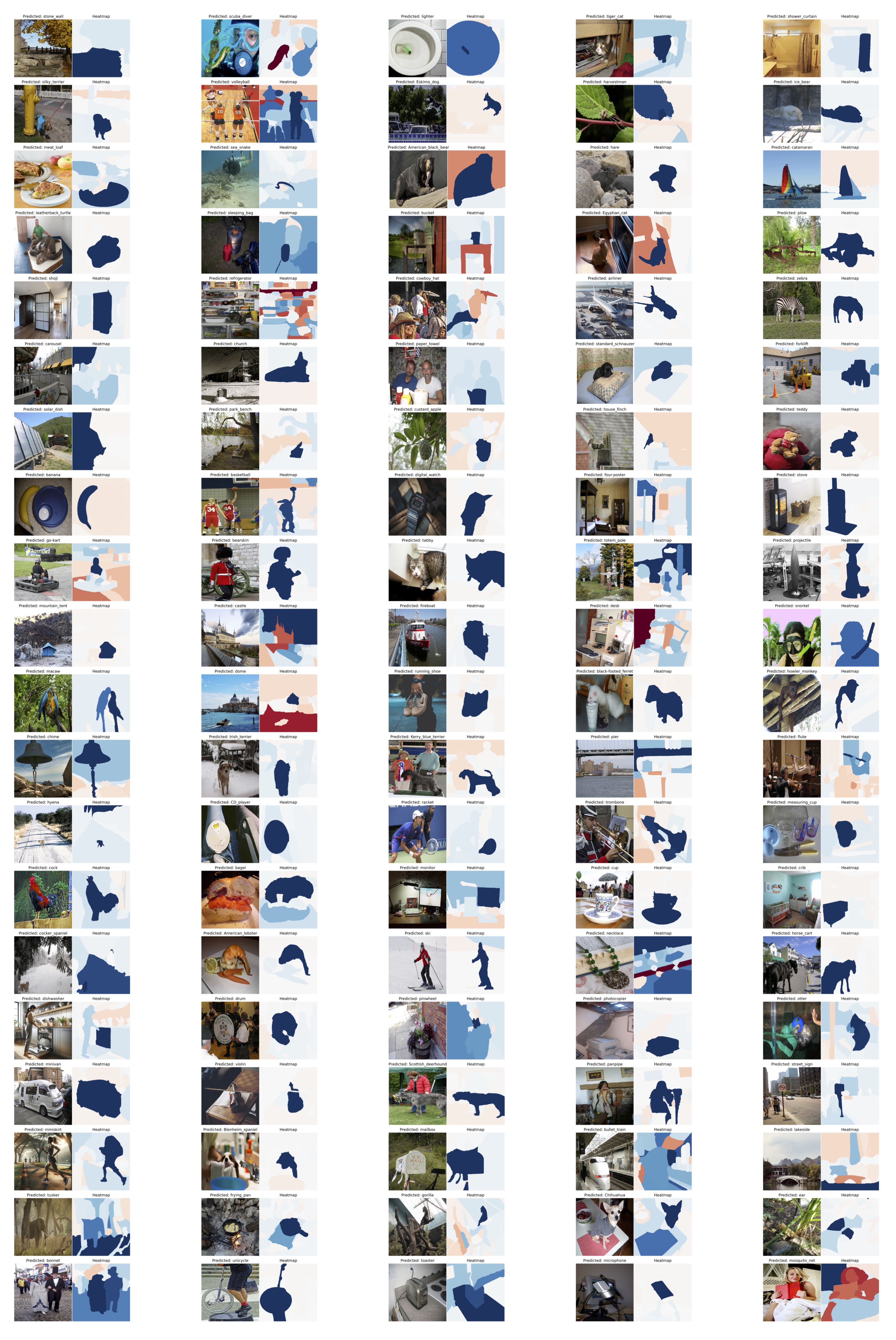}
\caption{\textbf{DSEG attribution maps.} Representation of the feature weights of all 100 images, with blue segments indicating positively important and red segments negatively important features in relation to the classified label.}
\label{heatmaps}
\end{figure}

In addition to visualizing the $n$ most essential segments for an explanation, feature attribution maps also help the explainee (the person receiving the explanation \citep{MILLER20191}) to get an idea of which other segments are important for interpreting the result. 
In these maps, the segments represent the corresponding coefficient of the surrogate model learned within LIME for the specific case. \Cref{heatmaps} represents all feature maps of EfficientNetB4 with the reported settings, accompanied by the original image with the most probable class. Blue segments are positively associated with the class to be explained, and red segments are negatively associated.

\newpage
\section{Dataset and user study} \label{us}

\subsection{Dataset} \label{app_data}

\textbf{Image selection.} As mentioned in \Cref{dataset}, we selected various classes of images from the ImageNet \citep{5206848} and COCO \citep{coco} dataset.
Additionally, we created artificial images using the text-to-image model DALL-E \citep{DBLP:journals/corr/abs-2102-12092} to challenge the XAI techniques when facing multiple objects. The dataset for the evaluation comprised 97 real images and three synthetic images. For the synthetic instances, the prompts 'realistic airplane at the airport' (Airplane), 'realistic person running in the park' (Miniskirt), and 'realistic person in the kitchen in front of a dishwasher' (Dishwasher) were used. 

The object types listed in \Cref{classes} represent the primary labels of the images used in the dataset. Each image is unique, ensuring no duplication and maximizing the diversity of animals and objects covered. The bolded types denote those that were randomly selected for qualitative evaluation. The cursive types signify the reduced dataset comprising fifty instances, implemented to conserve computational time for supplementary evaluations. These types provide a balanced representation of the dataset and were chosen to ensure broad coverage across different categories. This selection strategy helps to avoid bias and supports a comprehensive evaluation.

\begin{table}[ht]
 \caption{\textbf{Object families and types.} This table categorizes the images in the dataset according to their object families. The bold text denotes the classes selected for the user study, while the italicized text represents the minimized dataset; both were chosen at random to ensure objectivity variety.}
\vspace{0.2cm}
\centering
\label{classes}
\resizebox{\columnwidth}{!}{ 
\Large
\begin{tabular}{l|l}
\toprule
\textbf{Object Family} & \textbf{Type} \\ \midrule

\multirow{3}{*}{Animals} 
    & \textit{\textbf{Ice\_bear, Gorilla, Chihuahua, Husky, Horse}, Irish\_terrier, Macaw,}\\ 
    & \textit{American\_lobster, Kerryblue\_terrier, Zebra, House\_finch, American\_egret,}\\ 
    & \textit{Little\_blue\_heron, Tabby, Black\_bear, Egyptian\_cat}, Tusker, Quail, Affenpinscher\\ 
    & Leatherback\_turtle, Footed\_ferret, Howler\_monkey, Blenheim\_spaniel, Otter\\ 
    & Silky\_terrier, Cocker\_spaniel, Hare, Siberian\_husky, Harvestman, Sea\_snake \\
    &Cock, Scottish\_deerhound, Tiger\_cat, Hyena\\
    
    \midrule

\multirow{5}{*}{Objects} 
    & \textit{\textbf{Street\_sign, Park\_bench, CD\_player, Banana, Projectile, Ski,}} \\
    & \textit{\textbf{Catamaran, Paper\_towel, Violin, Miniskirt, Basketball, Tennis\_racket,}} \\
    & \textit{\textbf{Airplane, Dishwasher, Scuba\_diver}, Pier, Mountain\_tent, Totem\_pole,}\\ 
    & \textit{Bullet\_train, Lakeside, Desk, Castle, Running\_shoes, Snorkel, Digital\_Watch,} \\
    & \textit{Church, Refrigerator, Meat\_loaf, Dome, Forklift, Teddy, Mosque, Shower\_curtain}\\
    &Four\_poster, Photocopier, Stone\_wall, Crib, Bow\_tie, Measuring\_cup, Unicycle \\
    & Cowboy\_hat, Dutch\_oven, Go-kart, Necklace, Bearskin, Sleeping\_bag, Trombone\\
    & Microphone, Sandal, Fireboat, Carousel, Drum, Shoji, Solar\_dish, Stove, Cup\\
    & Panpipe, Custard\_apple, Gondola, Minivan, Bagel, Lighter, Pot, Carton, Ear\\
    & Volleyball, Plow, Mailbox, Bucket, Chime, Toaster \\

\bottomrule
\end{tabular}%
}
\end{table}

\subsection{User study} \label{user_appendix}

We conducted our research and user study using MTurk, intentionally selecting participants without specialized knowledge to ensure the classes represented everyday situations. Each participant received compensation of \$4.50 per survey, plus an additional \$2.08 handling fee charged by MTurk and \$1.24 tax. The survey, designed to assess a series of pictures, takes approximately 10 to 15 minutes to complete. The sequence in which the explanations are presented to the participants was randomized to minimize bias. In our study conducted via MTurk, 59 individuals participated, along with an additional 28 people located near our research group who participated at no cost.

\begin{figure}[h]
    \centering
    \begin{minipage}[t]{0.48\columnwidth}
        \centering
        \includegraphics[width=\linewidth]{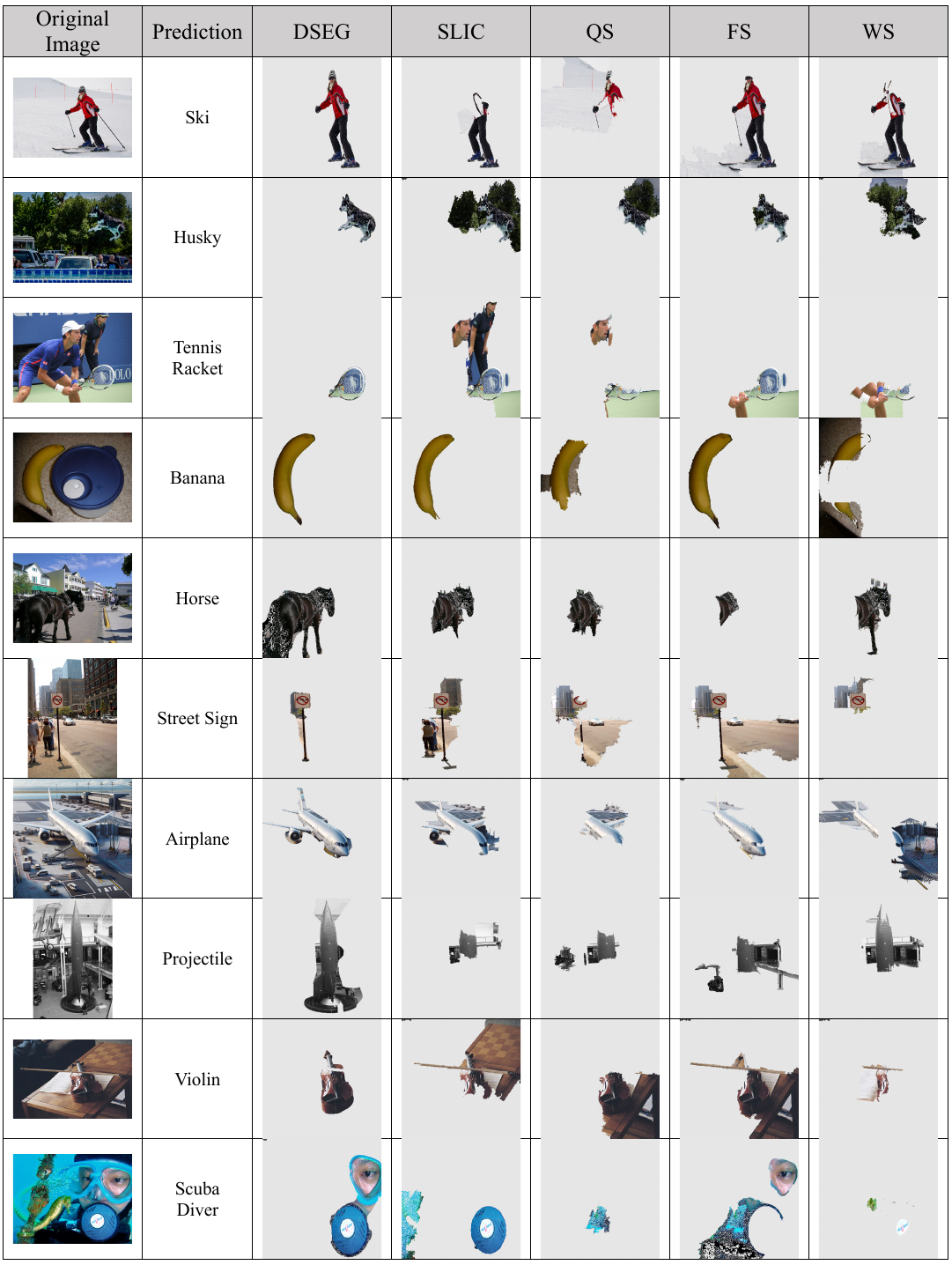}
        \caption{\textbf{Images 1–10.}}
        \label{fig:A1}
    \end{minipage}
    \hfill
    \begin{minipage}[t]{0.48\columnwidth}
        \centering
        \includegraphics[width=\linewidth]{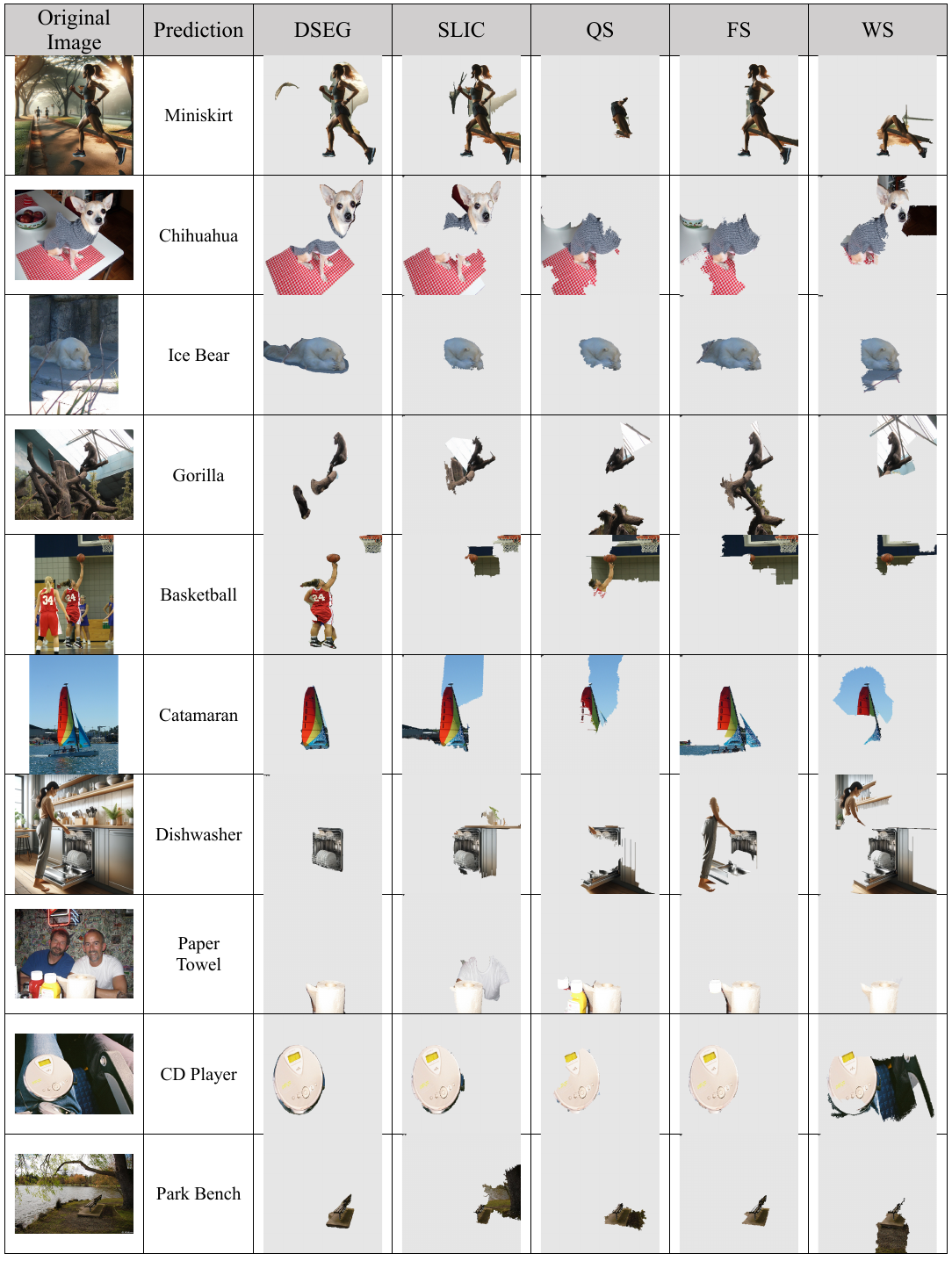}
        \caption{\textbf{Images 11–20.}}
        \label{fig:A2}
    \end{minipage}
    \caption{\textbf{User study data.} Examples from the evaluation datasets showing the LIME explanations alongside the original images and their corresponding predictions.}
    \label{fig:user_study_data}
\end{figure}

\newpage
\textbf{Explanations.}
In \Cref{fig:A1,fig:A2} we show all 20 images from the dataset used for the qualitative evaluation. Each image is accompanied by the prediction of EfficientNetB4 and the explanations within the vanilla LIME framework with all four segmentation approaches and the DSEG variant. The segments shown in the image indicate the positive features of the explanation. 

\begin{wrapfigure}[17]{r}{8cm} 
\vspace{-10pt} 
  \centering
  \includegraphics[width=0.5\linewidth]{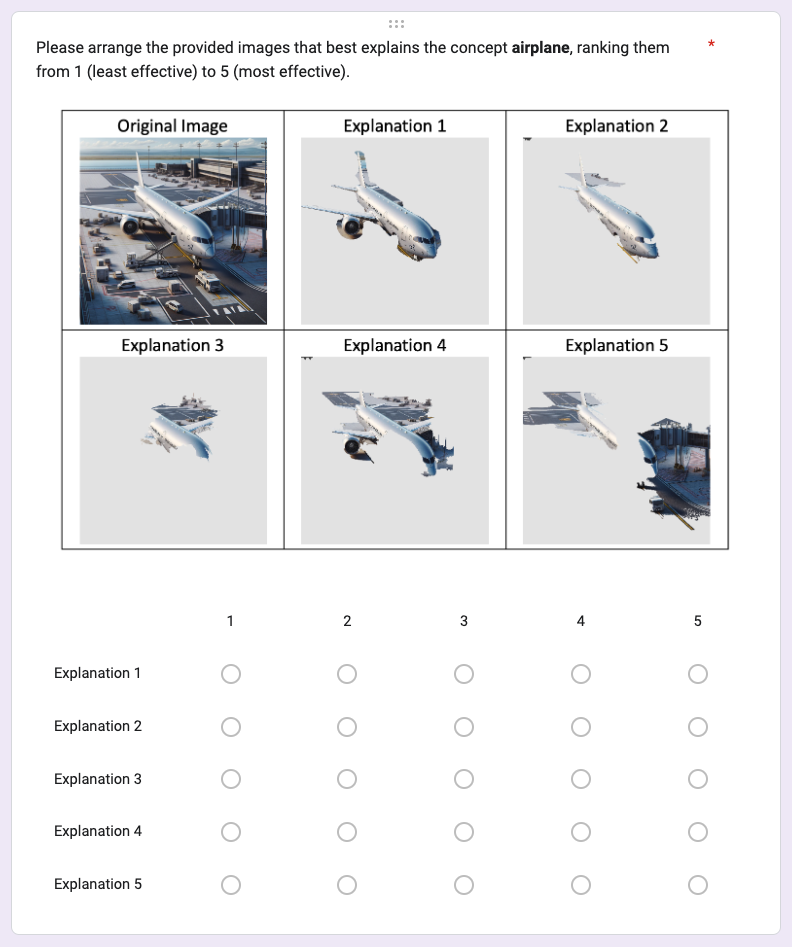}
\caption{\textbf{Exemplary question.} The 'airplane' example is shown in the original image with its five explanations. Below the images, participants can rate the quality of the explanations accordingly.}
\label{user_question}
\end{wrapfigure}

\textbf{Instruction.} Participants were tasked with the following question for each instance: 'Please arrange the provided images that best explain the concept [\textit{model's prediction}], ranking them from 1 (least effective) to 5 (most effective).' Each instance was accompanied by DSEG, SLIC, Watershed, Quickshift, Felzenszwalb, and Watershed within the vanilla LIME framework and the hyperparameters discussed in the experimental setup. These are also the resulting explanations used in the quantitative evaluation of EfficientNetB4. \Cref{user_question} shows an exemplary question of an instance of the user study.

\begin{figure}[ht]
     \centering
     \begin{minipage}[t]{0.48\textwidth}
         \centering
         \includegraphics[width=0.8\textwidth]{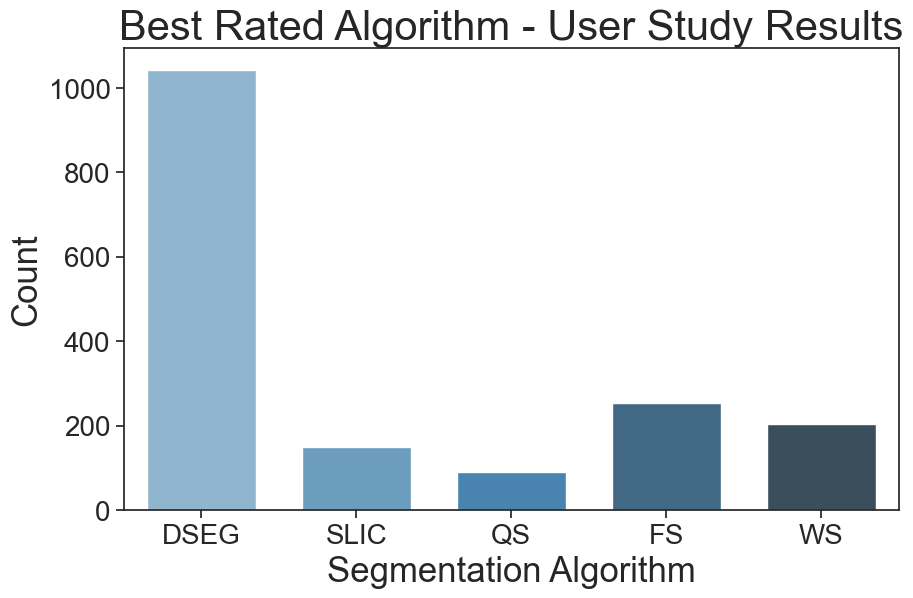}
         \caption{\textbf{Best rated explanation.} Accumulated number of best-selected explanations within the user study. DSEG was selected as the favourite, followed by Felzenszwalb and Watershed. }
         \label{asolute_rating}
     \end{minipage}
     \hfill
     \begin{minipage}[t]{0.48\textwidth}
         \centering
         \includegraphics[width=0.8\textwidth]{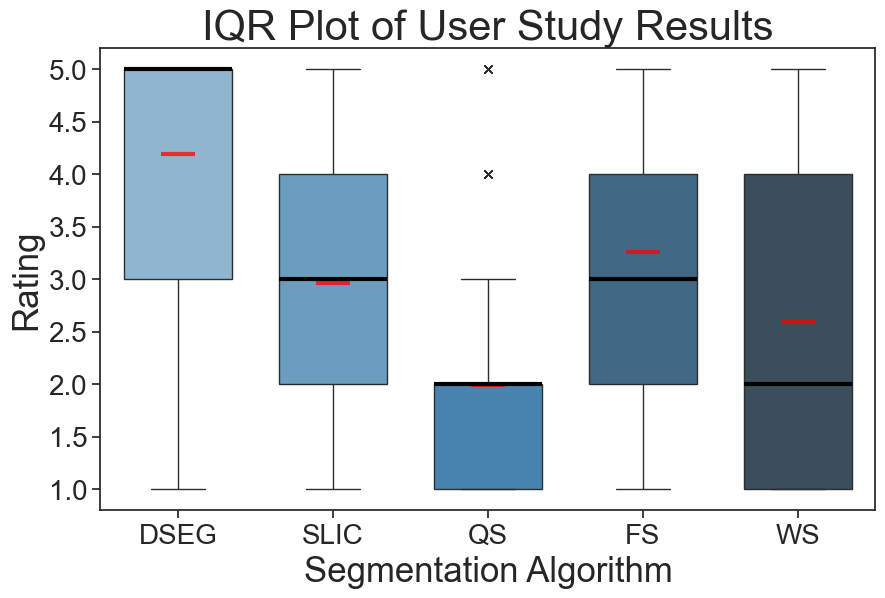}
         \caption{\textbf{IQR of explanation's ratings.} The IQR plot of the user study ratings is detailed, with the black line indicating the median and the red line representing the mean. This plot shows that DSEG received the highest ratings, while Watershed exhibited the broadest ratings distribution.}
         \label{iqr_rating}
     \end{minipage}
     \caption{\textbf{User study results.} The user study ratings are visualized in two distinct figures, each employing a different form of data representation. In both visualizations, DSEG consistently outperforms the other techniques.}
     \label{ratings}
\end{figure}

\textbf{Results.} We show the cumulative maximum ratings in \Cref{asolute_rating} and in \Cref{iqr_rating} the median (in black), the interquartile range (1.5), and the mean (in red) for each segmentation technique. DSEG stands out in the absolute ratings, significantly exceeding the others. Similarly, in \Cref{iqr_rating}, DSEG achieves the highest rating, indicating its superior performance relative to other explanations. Therefore, while DSEG is most frequently rated as the best, it consistently ranks high even when it is not the leading explanation, as the IQR of DSEG shows. Aligned with the quantitative results in \Cref{eval_section}, the Quickshift algorithm performs the worst. 

\newpage
\Cref{t_test} presents the statistical significance of the user study. Specifically, it lists the t-statistics and p-values for comparisons between DSEG (the baseline method) and other segmentation methods, namely SLIC, QS, FS, and WS. The t-statistics indicate the magnitude of difference between DSEG and each different method, with higher values representing greater differences. The corresponding p-values demonstrate the probability that these observed differences are due to random chance, with lower values indicating stronger statistical significance.

\begin{table}[h] 
\centering
\caption{\textbf{User study statistical results.} This table summarizes the statistical significance of user study results for each segmentation approach. The t-statistics and p-values indicate the comparison between DSEG and other methods. Extremely low p-values suggest strong statistical significance.}
\label{t_test}
\vskip 0.15in
\centering
\footnotesize
\begin{tabular}{l|rrrrr}
\toprule
Metric & DSEG & SLIC & QS & FS & WS \\
\midrule
t-statistics $\uparrow$ & -- & 20.01 & 49.39 & 20.89 & 33.15 \\
p-values $\downarrow$ & -- & 8.0e-143 & $<$ 2.2e-308 & 1.2e-86 & 3.3e-187 \\
\bottomrule
\end{tabular}

\end{table}

In this context, the null hypothesis (H$_0$) posits no significant difference in participant preferences between the performance of DSEG and other segmentation methods within the LIME framework. The alternative hypothesis (H$_A$) asserts that DSEG performs significantly better than the different segmentation methods on the dataset when evaluated using five explanations.
Given the extremely low p-values (e.g., 8.0e-143 for SLIC and < 2.2e-308 for QS), we can reject the null hypothesis (H$_0$) with high confidence. The significance level of 99.9\% ($\alpha$ = 0.001) further supports this conclusion, as all p-values fall well below this threshold. These results indicate that the observed differences are highly unlikely to have occurred by chance and are statistically significant.

\end{document}